\documentclass[10pt,twocolumn,letterpaper]{article}

\usepackage[T1]{fontenc}
\usepackage[utf8]{inputenc}
\usepackage{mathptmx}
\usepackage[scaled=0.92]{helvet}
\usepackage{courier}
\usepackage{amsmath,amssymb,amsthm,mathtools,bm}
\usepackage[margin=0.75in,columnsep=0.25in]{geometry}
\usepackage{microtype}
\usepackage{graphicx}
\usepackage{array,booktabs,multirow,makecell}
\usepackage{xcolor}
\usepackage[font=small,labelfont=bf]{caption}
\usepackage{enumitem}
\usepackage{tikz}
\usetikzlibrary{arrows.meta,positioning,calc,fit,backgrounds}
\usepackage[numbers,sort&compress]{natbib}
\usepackage{authblk}
\usepackage{url}
\usepackage[colorlinks=true,linkcolor=blue!50!black,citecolor=blue!50!black,urlcolor=blue!50!black]{hyperref}
\usepackage[capitalise,noabbrev]{cleveref}
\usepackage{xspace}

\setlist{leftmargin=*,itemsep=1pt,topsep=2pt}
\theoremstyle{plain}
\newtheorem{theorem}{Theorem}
\newtheorem{proposition}[theorem]{Proposition}
\newtheorem{corollary}[theorem]{Corollary}
\newtheorem{lemma}[theorem]{Lemma}
\newtheorem{conjecture}[theorem]{Conjecture}
\theoremstyle{definition}
\newtheorem{definition}[theorem]{Definition}
\theoremstyle{remark}
\newtheorem{remark}[theorem]{Remark}
\crefname{conjecture}{Conjecture}{Conjectures}

\newcommand{\R}{\mathbb{R}}
\newcommand{\N}{\mathbb{N}}
\newcommand{\Cset}{\mathcal{C}}
\newcommand{\Xs}{\mathcal{X}}
\newcommand{\id}{\mathrm{id}}

\newcommand{\ns}{\textsc{NS}\xspace}
\newcommand{\sel}{\textsc{Sel}\xspace}
\newcommand{\selm}{\textsc{Sel-Merge}\xspace}
\newcommand{\zero}{\textsc{Zero}\xspace}

\newcommand{\yord}{y_{\mathrm{ord}}}
\newcommand{\Jh}{\hat{J}}
\newcommand{\Lh}{\hat{\Lambda}}
\newcommand{\Phih}{\hat{\Phi}}
\DeclareMathOperator{\diag}{diag}

\title{\Large\bfseries Learning Without Clocks: Partition-Invariant Neural Networks}

\author[1]{Han Wang}
\author[1,2]{Jingyuan Bao}
\author[1,2]{Juntao Wu}
\author[1]{Saiao Zhou}
\author[1,*]{Yuman Nie}
\affil[1]{Hefei Institutes of Physical Science, Chinese Academy of Sciences, Hefei 230031, China}
\affil[2]{Science Island Branch, Graduate School of University of Science and Technology of China, Hefei 230026, China}
\affil[*]{Corresponding author: Yuman Nie}

\date{}

\begin{document}

\twocolumn[
\begin{@twocolumnfalse}
\maketitle
\vspace{-1.5em}
\begin{abstract}
\noindent
Event cameras, spiking cochleas and many other sensors report signals as asynchronous, time-stamped events, yet most learning systems first impose a clock: events are binned into frames or processed in packets, and the output then depends on how the stream was sliced. We study networks whose state is updated exactly at every event and evolves in closed form between events. Such models are invariant to how a stream is partitioned into packets, a property they share with recent signature-based models of irregular data. We ask what more is needed for the output to be (i)~independent of the processing order of simultaneous events and (ii)~continuous in the time stamps, including when two events cross in time, while (iii)~remaining sensitive to temporal order. For arbitrary nonlinear event models we prove that (i) and (ii) are each equivalent to pairwise commutation of the event updates, and that once the updates commute, every order-dependent effect originates in the non-commutation between event updates and the inter-event dynamics and vanishes linearly with the time lag. This identifies a class of \emph{time-mediated non-commutative} (TM) models and yields a resolution--robustness bound that holds for any model. We construct the minimal TM layer, a learnable and provably Lipschitz generalization of the Hassenstein--Reichardt correlator; prove that no single additive layer, whatever its readout, can uniformly represent ordered pair interactions, whereas a three-dimensional TM state can; show that each TM layer contributes two levels of temporal nesting against one for additive layers; and give a single-layer higher-order construction (TM-$k$) and a stacking rule that preserve all three properties. We instantiate these results as the Clock-Free Event (CFE) layer, whose cost is linear in the number of events, describe a topology-free variant for temporal point processes with closed-form likelihoods, and specify a preregistered evaluation protocol for clock robustness, including a physical-time threat model for spike-retiming attacks. In preliminary development-set experiments, the ordered TM state improves on an otherwise identical additive state by 3.5--9.1 percentage points on DVS128 Gesture, depending on the observation window, and by 24.5 and 30.6 points on Spiking Heidelberg Digits and Speech Commands.
\end{abstract}
\vspace{1em}
\end{@twocolumnfalse}
]

\section{Introduction}
\label{sec:intro}

\begin{figure*}[t]
\centering
\includegraphics[width=\textwidth]{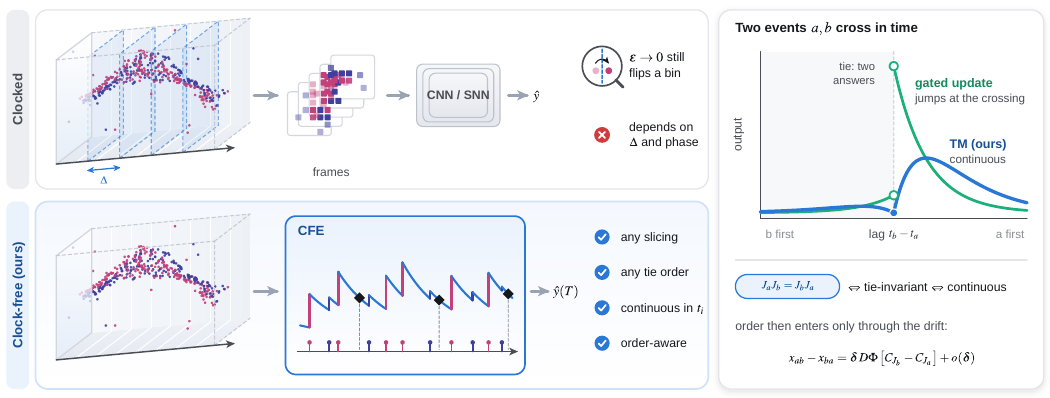}
\caption{Clocked versus clock-free processing. \emph{Top:} binned pipelines slice the event stream into frames of width $\Delta$; the input tensor depends on $\Delta$ and on the phase of the grid, and moving one event across a bin edge by an arbitrarily small $\varepsilon$ changes a count (\cref{prop:binned}). \emph{Bottom:} a CFE network updates its state exactly at every event (jumps, coloured by event type) and evolves it in closed form between events (blue); queries ($\blacklozenge$) can be placed at arbitrary times, and the result does not depend on how the stream is cut into packets (\cref{prop:homomorphism}). \emph{Right:} output of a single site as two events of types $a$ and $b$ cross in time, computed from the model equations with illustrative parameters. A gated update (\sel) jumps at the crossing and returns two different values at an exact tie, depending on the processing order; the TM layer is continuous and still distinguishes the two orders. Tie invariance and crossing continuity are each equivalent to commuting jumps (\cref{thm:commute}), and order information then enters only through the non-commutation of the jumps with the drift (\cref{thm:timemediated}).}
\label{fig:overview}
\end{figure*}

Many sensors do not sample the world on a clock. A dynamic vision sensor emits an event whenever the log intensity at a pixel changes by a threshold~\cite{lichtsteiner2008,gallego2022survey}; a silicon cochlea emits spikes on frequency channels~\cite{cramer2022shd}; neural recordings, user activity logs and clinical measurements arrive at irregular, data-dependent times. In such signals much of the information is carried by \emph{when} events occur and in \emph{which order}, for instance the direction of a moving edge or the order of phonetic onsets.

Most learning pipelines nevertheless begin by imposing a clock. Events are binned into frames or voxel grids and processed by convolutional, recurrent or spiking networks with a fixed time step~\cite{fang2021plif,amir2017gesture}, or they are grouped into packets of fixed duration or fixed count. The output then depends on the bin width, the phase of the bin grid and the packet boundaries, quantities that are artifacts of the pipeline rather than properties of the signal. Hard binning is discontinuous in the time stamps~\cite{chowdhury2026tradeoff}, recurrent and transformer detectors trained at one frequency degrade sharply at others~\cite{zubic2024ssm}, and spiking networks can be fooled by moving a small fraction of spikes by a single time step~\cite{yu2026retiming}.

A growing line of work processes events asynchronously, one at a time~\cite{messikommer2020asynet,schaefer2022aegnn,sekikawa2019eventnet,schoene2024eventssm,hao2026ssla,pellerito2026neuralevents}, and continuous-time models driven by rough-path signatures compose their state over arbitrary intervals~\cite{walker2026faithful,bloch2026ews}. When the state is updated exactly at each event and evolves by an exact semigroup between events, the output is \emph{partition invariant}: it does not depend on how the stream is cut into packets (\cref{prop:homomorphism}; see~\cite{walker2026faithful} for the corresponding interval-composition property of online log-signature models). Partition invariance is the principle behind this paper's title: the output should depend on the signal, not on the clock used to sample it or on how it is sliced.

Partition invariance alone does not make a model robust. Real sensors produce \emph{simultaneous} events, because time stamps are quantized (often to 1\,\textmu s), so the processing order within a tie is arbitrary. Time stamps also jitter, so the output should vary continuously with them, including at configurations where two events \emph{cross} in time. Yet temporal order must remain visible to the model. These requirements appear to be in tension: an update rule that is insensitive to the order of simultaneous events might seem unable to detect order at all (\cref{fig:overview}).

\paragraph{Results.} We resolve this tension with a short chain of theorems that hold for general nonlinear event models.
\begin{enumerate}
\item \textbf{Ties, crossings and commutation} (\cref{thm:commute}). For any event model, invariance to the processing order of two simultaneous events, continuity of the state when the two events cross in time, and commutation of the two event updates are equivalent. Consequently, when updates do not commute, no rule for resolving exact ties (sequential, group-wise or Marcus-type) can restore continuity (\cref{cor:norescue}). Selective or gated updates of the kind used in modern recurrent and state-space models therefore make the output discontinuous at every crossing of event types whose updates fail to commute.
\item \textbf{Time-mediated order} (\cref{thm:timemediated}). Once updates commute, every order-dependent effect originates in the non-commutation between the event updates and the inter-event dynamics, and it vanishes linearly with the time lag. Models whose updates also commute with the dynamics are blind to both time and order. This characterizes the class of \emph{time-mediated non-commutative} (TM) models, and yields a resolution--robustness bound valid for any model (\cref{cor:resolution}): resolving the order of two events at lag $\delta$ with margin $m$ requires a timing sensitivity of at least $m/(2\delta)$.
\item \textbf{A minimal TM layer and its expressivity} (\cref{sec:tm}). Writing events into a fast trace $u$, letting the dynamics copy $u$ into a delayed trace $s$, and letting events read $s$ into an accumulator $v$ gives the smallest TM layer with ordered pair interactions. Its state is an exponentially weighted sum over ordered event pairs, a learnable form of the Hassenstein--Reichardt correlator~\cite{hassenstein1956}, and it is Lipschitz in the time stamps (\cref{thm:lipschitz}). We prove that a single additive layer cannot represent ordered pair interactions uniformly in the number of events, whatever its width and readout, whereas three TM state dimensions suffice (\cref{thm:separation,prop:tmyord}). We also show that each TM layer adds two levels of temporal nesting against one for additive layers (\cref{prop:height}), and we give a single-layer construction of arbitrary order (TM-$k$, \cref{prop:tmk}) and a stacking rule that preserves all robustness properties (\cref{prop:stacking}).
\item \textbf{The Clock-Free Event (CFE) layer} (\cref{sec:cfe,fig:cfe}). We instantiate TM models with local sites and shared letters for sensors with topology, and with type sites for event sequences without topology. The layer is computed exactly, event by event or with parallel scans, at a cost linear in the number of events. We also give a send-on-delta encoder that turns sampled signals into events consistently across sampling rates, and intensity heads for temporal point processes whose likelihood is available in closed form.
\item \textbf{An evaluation protocol for clock robustness} (\cref{sec:protocol}). We specify a preregistered protocol with single-use test sets, clock-shift conditions applied without retraining, compute-matched baselines, and a physical-time threat model for spike-retiming attacks that follows from the theory.
\end{enumerate}

\paragraph{Scope of this version.} This version reports the theory, the architecture, the evaluation protocol and preliminary development-set experiments (\cref{sec:experiments}). In those experiments the ordered TM state improves on an otherwise identical additive state by 3.5--9.1 percentage points on DVS128 Gesture~\cite{amir2017gesture}, depending on the observation window, and by 24.5 and 30.6 points on Spiking Heidelberg Digits (SHD)~\cite{cramer2022shd} and Speech Commands~\cite{warden2018speech}. Official test-set results under the preregistered protocol will be reported in a later version. We do not claim state-of-the-art accuracy; \cref{sec:discussion} discusses the remaining gaps.

\section{Related Work}
\label{sec:related}

\paragraph{Asynchronous event processing.} Event-based vision is surveyed in~\cite{gallego2022survey}. Several methods update a representation per event in order to save computation. AsyNet makes sparse convolutions asynchronous~\cite{messikommer2020asynet}, AEGNN updates graph networks locally~\cite{schaefer2022aegnn}, EventNet processes events recursively~\cite{sekikawa2019eventnet}, and SSLA restricts a linear-attention state to spatial substates activated by each event~\cite{hao2026ssla}. Neural Events learns a discrete asynchronous event code~\cite{pellerito2026neuralevents}, and Event2Vec embeds events directly as vectors~\cite{fang2026event2vec}. Event-SSM processes events with deep state-space layers and exact inter-event decay~\cite{schoene2024eventssm}; its formulation assumes strictly increasing time stamps, applies a nonlinearity to every event output and pools over consecutive event indices. Zubi\'c et al.\ showed that SSM-based detectors trained at one input frequency generalize to others far better than recurrent and transformer detectors~\cite{zubic2024ssm}. In the works we surveyed, the behavior of the model under ties and under crossings of event times is not characterized; this is the subject of \cref{sec:theory}.

\paragraph{Spiking networks and timing.} Spiking networks with learnable membrane time constants (PLIF)~\cite{fang2021plif}, implemented in SpikingJelly~\cite{fang2023spikingjelly}, are standard baselines for event data. They are usually simulated with a fixed time step. EventProp computes exact gradients for event-driven spiking networks~\cite{wunderlich2021eventprop}, learnable delays improve audio benchmarks~\cite{hammouamri2024dcls}, Bullet Trains parallelizes exact event-driven simulation with associative scans~\cite{morrill2026bullet}, and EGRU exploits activity sparsity in recurrent networks~\cite{subramoney2023egru}. On the robustness side, hard temporal binning has no finite Lipschitz constant and continuous binning encoders obey a $2/\Delta$ sensitivity floor~\cite{chowdhury2026tradeoff}, and timing-only attacks that retime a small fraction of spikes defeat spiking networks while existing defenses help little~\cite{yu2026retiming}. We give the order-resolution counterpart of the binning floor (\cref{cor:resolution}) and a threat model in physical time (\cref{sec:protocol}).

\paragraph{Continuous-time and state-space sequence models.} ODE-RNNs~\cite{rubanova2019latentode} and GRU-ODE-Bayes~\cite{debrouwer2019gruodebayes} evolve a hidden state between irregular observations and update it with gated maps at observations. Neural controlled differential equations~\cite{kidger2020ncde} and their log-signature variants~\cite{walker2024logncde} drive a vector field by an interpolated path. Structured state-space models~\cite{gu2022s4,smith2023s5} and selective variants~\cite{gu2023mamba} are linear in the hidden state. Muca Cirone et al.\ characterized the expressivity of selective models through the signature, showing that diagonal transitions only compare pairs of time points while dense ones see whole segments~\cite{cirone2024selective}; structured linear CDEs recover maximal expressivity at lower cost~\cite{walker2025slice}. The exponentially weighted signature extends SSMs with matrix-exponential weighting of iterated integrals~\cite{bloch2026ews}. Closest to our starting point, Walker et al.\ embed irregular and asynchronous observations as event increments with count channels and compose truncated log-signatures over arbitrary query intervals~\cite{walker2026faithful}. Their formalism excludes tied observation times and establishes continuity within a fixed ordering of events, and their experiments use non-commuting generators; the tensor dimension grows as $d^N$ with the number of channels $d$ and truncation level $N$. Our results concern exactly the regime this leaves open: ties, crossings of event times, and high-dimensional sensors. The question of whether recurrent networks can warp time~\cite{tallec2018warp} is related in spirit but concerns reparameterizations rather than slicing.

\paragraph{Rough paths with jumps.} Marcus' canonical interpretation~\cite{marcus1981} and its rough-path generalization~\cite{chevyrev2019canonical} handle discontinuous drivers. Chevyrev and Friz note that when jumps in different directions approach each other, the limiting path function depends on their order (their Example~3.20); \cref{thm:commute} turns this phenomenon into a characterization for learning models. Universal approximation for functionals of c\`adl\`ag paths is studied in~\cite{cuchiero2025cadlag}. Chen's identity~\cite{chen1957} and rough path theory~\cite{lyons1998} underlie interval composition; commutation relations among letters, as in trace monoids~\cite{cartier1969}, reappear in our characterization.

\paragraph{Motion detection and sampling-rate independence.} The Hassenstein--Reichardt correlator multiplies a delayed signal from one receptor with the signal of a neighbor~\cite{hassenstein1956}; the TM layer is a learnable, multichannel, provably smooth generalization (\cref{prop:closedform}). Sampling-frequency-independent convolution layers make audio networks independent of the sampling rate by parameterizing analog filters~\cite{saito2021sfi}. Our approach reaches the same goal by consuming events in continuous time (\cref{sec:encoder}) and extends it to slicing, jitter and ties.

\section{Event Models: Partitions, Ties and Crossings}
\label{sec:theory}

\subsection{Setting}

\begin{definition}[Event stream]
An event stream on a window starting at $t_0$ is a finite sequence $S=((t_1,c_1),\dots,(t_n,c_n))$ with $t_0\le t_1\le\dots\le t_n$ and \emph{types} $c_i\in\Cset$. A type collects everything that is intrinsic to one event: sensor location or channel, polarity, and any \emph{marks} computed from that event alone. Events with equal time stamps form a \emph{tie group}.
\end{definition}

\begin{definition}[Event model]
An event model $\mathcal{M}=(\Xs,\Phi,(J_c)_{c\in\Cset},x_0)$ consists of a state space $\Xs$ (a metric space), a continuous semiflow $\Phi:[0,\infty)\times\Xs\to\Xs$ with $\Phi_0=\id$ and $\Phi_{\delta'}\circ\Phi_\delta=\Phi_{\delta+\delta'}$ (the \emph{drift}), continuous \emph{jump} maps $J_c:\Xs\to\Xs$, and an initial state $x_0$. Given an ordering of the events that is consistent with the time stamps, the state at a query time $T\ge t_0$ is
\begin{equation}
x(T)=\Phi_{T-t_m}\circ J_{c_m}\circ\Phi_{t_m-t_{m-1}}\circ\cdots\circ J_{c_1}\circ\Phi_{t_1-t_0}(x_0),
\label{eq:state}
\end{equation}
where $m=\max\{i: t_i\le T\}$, so that a query at $T$ includes every event with $t_i\le T$. A readout maps the states at one or more query times to outputs.
\end{definition}

This definition covers additive and selective linear state-space layers with exact inter-event decay, ODE-RNN-type models with gated updates at observations (when the drift is integrated exactly), and the layers introduced in \cref{sec:tm}. For affine models on $\Xs=\R^d$ we write $J_c(x)=G_cx+b_c$ with $G_c=I+N_c$ and $\Phi_\delta=e^{\delta\Lambda}$.

\subsection{Partition invariance}

A \emph{packet} is a half-open interval $(a,b]$ together with the events it contains, so packets never split a tie group. A packet processor assigns to each packet a map $P(a,b]:\Xs\to\Xs$; processing a stream with cut points $a=\tau_0<\tau_1<\dots<\tau_K=b$ produces $P(\tau_{K-1},\tau_K]\circ\cdots\circ P(\tau_0,\tau_1]$. The processor is \emph{partition invariant} if the result does not depend on the cut points.

\begin{proposition}[Partition invariance is a homomorphism property]
\label{prop:homomorphism}
(a) A packet processor is partition invariant if and only if $P(a,c]=P(b,c]\circ P(a,b]$ for all $a<b<c$. (b) Exact processing of an event model, $P(a,b]=\Phi_{b-t_m}\circ J_{c_m}\circ\cdots\circ J_{c_\ell}\circ\Phi_{t_\ell-a}$ over the events in $(a,b]$, is partition invariant. (c) If the processor is time-homogeneous, maps empty packets to linear maps on $\R^d$ that depend continuously on the packet length, and maps empty packets of length zero to the identity, then the empty-packet maps form a matrix semigroup $e^{\delta\Lambda}$.
\end{proposition}

Part (a) is the composition law of Chen's identity~\cite{chen1957} transported to packets; the interval-composition property of online log-signature models~\cite{walker2026faithful} and the chunked exact simulation of spiking networks~\cite{morrill2026bullet} are instances of it. Part (c) rules out per-packet nonlinear updates and clock-dependent dynamics, which is why frame-based and packet-based recurrent pipelines are not partition invariant. Proofs are in \cref{app:proofs}. A readout that keeps a memory across query times makes the network output depend on the \emph{query schedule}; it remains independent of the partition, and we keep the two notions separate throughout.

\subsection{Ties and crossings}

\begin{definition}
Let $a,b\in\Cset$. The model is \emph{tie invariant} for $(a,b)$ if $J_b(J_a(x))=J_a(J_b(x))$ for every state $x$ reachable at a time when an $a$-event and a $b$-event can coincide; that is, the order in which a tie is processed does not matter. It is \emph{crossing continuous} for $(a,b)$ if, for every such $x$,
\[
\lim_{\delta\downarrow0}J_b\circ\Phi_\delta\circ J_a(x)=\lim_{\delta\downarrow0}J_a\circ\Phi_\delta\circ J_b(x),
\]
so that the state does not jump when the time stamps of an $a$-event and a $b$-event move past each other.
\end{definition}

\begin{theorem}[Ties, crossings and commutation]
\label{thm:commute}
For any event model and any pair of types $(a,b)$, the following are equivalent: (i) tie invariance; (ii) crossing continuity; (iii) $J_a\circ J_b=J_b\circ J_a$ on the reachable states. If (iii) holds for all pairs, the state $x(T)$ is independent of the processing order of every tie group, and it is a continuous function of the vector of time stamps on the whole set $\{t_0\le t_1,\dots,t_n\le T\}$ of configurations whose events do not cross $T$, not only within regions of fixed event order.
\end{theorem}

The proof uses only the continuity of $\Phi$ and of the jump maps, so the theorem applies to nonlinear models without modification (\cref{app:proofs}).

\begin{corollary}[No tie rule rescues non-commuting updates]
\label{cor:norescue}
If $J_aJ_b(x)\neq J_bJ_a(x)$ for some reachable $x$, then any rule that assigns a state to the exact tie leaves the state discontinuous at the crossing. This applies to processing ties in a canonical order, to group (Jacobi-type) updates that apply all jumps of a tie to the pre-tie state, and to Marcus-type rules that replace the tie by $\exp(A_a+A_b)$. Training with timing jitter can reduce the size of the jump only by driving the learned parameters toward commutation.
\end{corollary}

\begin{corollary}[Affine criterion]
\label{cor:affine}
Affine jumps $J_c(x)=G_cx+b_c$ commute if and only if $G_aG_b=G_bG_a$ and $N_ab_b=N_bb_a$, where $N_c=G_c-I$. In particular, translations ($N_c=0$) always commute; and updates of the form $N_c=g_cR_cE_{\mathrm{rd}}$ that \emph{read} a block of coordinates through a projection $E_{\mathrm{rd}}$ and \emph{write} into a different block commute whenever no update writes into a block that some update reads and the constant parts $b_c$ do not touch the read blocks. Selective updates that read the coordinates written by $b_c$, such as $v\leftarrow v+g_cR_cu$ with $u\leftarrow u+\alpha_c$, give $N_ab_b=g_aR_a\alpha_b\neq g_bR_b\alpha_a$ for generic parameters.
\end{corollary}

\begin{table*}[t]
\centering
\caption{Properties of event-model families under exact computation, for generic parameters. (P) partition invariance; (T) tie invariance; (C) continuity in the time stamps across crossings; (O$_2$) ordered pair interactions within one layer. Justifications are in \cref{app:taxonomy}.}
\label{tab:taxonomy}
\small
\begin{tabular}{@{}>{\raggedright\arraybackslash}p{0.31\textwidth}>{\raggedright\arraybackslash}p{0.25\textwidth}ccc>{\raggedright\arraybackslash}p{0.15\textwidth}@{}}
\toprule
Family & Update at an event & (P) & (T) & (C) & (O$_2$) \\
\midrule
Binned networks (frames, fixed time step; e.g.\ PLIF~\cite{fang2021plif}) & counts added to the current bin & no & yes & no & yes, at bin resolution \\
Additive event SSMs (exact decay; \ns) & $x\leftarrow x+b_c$ & yes & yes & yes & no (\cref{thm:separation}) \\
Selective or input-gated affine updates (\sel) & $x\leftarrow G_cx+b_c$, generic $G_c$ & yes & no & no & yes \\
Gated nonlinear updates with ODE drift~\cite{rubanova2019latentode,debrouwer2019gruodebayes} & $x\leftarrow \mathrm{GRU}(x,c)$ & yes & no & no & yes \\
Online log-signature models~\cite{walker2026faithful} & interval composition, non-commuting generators & yes & excl.$^\dagger$ & fixed order$^\ddagger$ & yes \\
\textbf{TM (this work)} & $u\leftarrow u+\alpha_c$, $v\leftarrow v+g_cR_cs$ & \textbf{yes} & \textbf{yes} & \textbf{yes} & \textbf{yes} \\
\bottomrule
\end{tabular}

\smallskip
\footnotesize $^\dagger$Ties are excluded by assumption. $^\ddagger$Continuity is established within a fixed order of events.
\end{table*}

\Cref{tab:taxonomy} applies these results to common model families. Ties are not exotic: sensors quantize time stamps (to 1\,\textmu s for the event recordings of \cref{sec:experiments}), and at high event rates distinct events share a time stamp, so in sequential models the arbitrary within-tie order becomes part of the input.

\subsection{Time-mediated order}

Commuting updates might seem to forbid order sensitivity. They do not, because the drift does not have to commute with the jumps.

\begin{theorem}[Time-mediation principle]
\label{thm:timemediated}
Let $\Xs=\R^d$, let $\Phi$ be the flow of a $C^1$ vector field $F$, and let the jump maps be $C^1$ and pairwise commuting. Define the \emph{flow commutator} of a jump, $C_J(y)=DJ(y)F(y)-F(J(y))$.
\begin{enumerate}[label=(\alph*)]
\item If every jump commutes with the flow, $J_c\circ\Phi_\delta=\Phi_\delta\circ J_c$ for all $\delta\ge0$ (equivalently $C_{J_c}\equiv0$), then $x(T)=\Phi_{T-t_0}\bigl(J_{c_m}\circ\cdots\circ J_{c_1}(x_0)\bigr)$: the state does not depend on any time stamp or on the order of events.
\item Consider two streams that coincide except that the events at times $t$ and $t+\delta$ have types $(a,b)$ in one and $(b,a)$ in the other, no other event occurs in $(t,t+\delta]$, and $Y$ is their common state just before $t$. Then
\begin{equation}
\begin{split}
x_{ab}(T)-x_{ba}(T)&=\delta\,D\Phi_{T-t}(J_aJ_bY)\\
&\quad\times\bigl[C_{J_b}(J_aY)-C_{J_a}(J_bY)\bigr]+o(\delta).
\end{split}
\label{eq:firstorder}
\end{equation}
\end{enumerate}
\end{theorem}

For affine models the expansion is exact. In homogeneous coordinates, with $\Jh_c=\bigl(\begin{smallmatrix}G_c&b_c\\0&1\end{smallmatrix}\bigr)$ and $\Lh=\bigl(\begin{smallmatrix}\Lambda&0\\0&0\end{smallmatrix}\bigr)$, the flow commutator is the matrix commutator $[\Jh_c,\Lh]$, and
\begin{equation}
x_{ab}(T)-x_{ba}(T)=\Phih_{T-t-\delta}\int_0^\delta\bigl(\Jh_b\Lh\Phih_\sigma\Jh_a-\Jh_a\Lh\Phih_\sigma\Jh_b\bigr)\,d\sigma\;\hat Y,
\label{eq:affineexact}
\end{equation}
whose leading term is $\delta\,\Phih_{T-t}\bigl([\Jh_b,\Lh]\Jh_a-[\Jh_a,\Lh]\Jh_b\bigr)\hat Y$ (\cref{app:proofs}).

\begin{corollary}[The time-mediated class]
\label{cor:tmclass}
An event model with commuting jumps is tie invariant and crossing continuous. Its sensitivity to time and to order is carried entirely by the flow commutators: it vanishes if all jumps commute with the drift, and otherwise the effect of swapping two events at lag $\delta$ is $O(\delta)$ with the coefficient in~\eqref{eq:firstorder}. We call models with pairwise commuting jumps and at least one jump that does not commute with the drift \emph{time-mediated non-commutative} (TM). For affine models with exact processing, (P), (T) and (C) hold if and only if the jumps commute, and the state then depends on time only through the commutators $[\Jh_c,\Lh]$.
\end{corollary}

Within the TM class, additive models (\ns) are sensitive to time and order only to first order: they know which type occurred more recently, but they cannot form interactions between pairs of events (\cref{thm:separation}). The TM layer of \cref{sec:tm} is the smallest member that forms ordered pair interactions.

\subsection{Resolution versus robustness}

\begin{corollary}[Resolution--robustness bound]
\label{cor:resolution}
Let a model's output $y$ satisfy $|y(t)-y(t')|\le L\|t-t'\|_1$ for time-stamp vectors $t,t'$. Then swapping the types of two events at lag $\delta$ changes $y$ by at most $2L\delta$. Distinguishing the two orders with margin $m$ therefore requires $\delta\ge m/(2L)$. A model that distinguishes them with a fixed margin at arbitrarily small lags is discontinuous.
\end{corollary}

Non-commuting updates (\cref{cor:norescue}) and binned inputs both buy unlimited order resolution with discontinuity. TM models trade it for continuity: their resolution is set by learnable time constants, and their sensitivity is bounded (\cref{thm:lipschitz}). The bound complements the $2/\Delta$ floor for continuous binning encoders~\cite{chowdhury2026tradeoff}.

\begin{proposition}[Binned inputs are discontinuous in physical time]
\label{prop:binned}
Consider any binning of time with bin width $\Delta$ and any phase. For every $0<\varepsilon\le\Delta/2$, each event whose time stamp lies within $\varepsilon$ of a bin edge can be moved into the adjacent bin by a retiming of size at most $\varepsilon$, which changes the binned input by $2$ in $\ell_1$, independently of $\varepsilon$. If time stamps are spread uniformly within bins, a fraction $2\varepsilon/\Delta$ of the events is exposed in this way. For a TM layer, a retiming of $k$ events by at most $\varepsilon$ each changes the state by at most $Ck\varepsilon$, where $C$ is the constant of \cref{thm:lipschitz}.
\end{proposition}

Under a physical-time budget, then, an adversary facing a binned model can choose events near bin edges and obtain one-bin moves at arbitrarily small cost, whereas against a TM layer the damage is proportional to the total displacement. This motivates the threat model of \cref{sec:protocol}.

\section{The Minimal TM Layer and Its Expressivity}
\label{sec:tm}

\subsection{Definition}

A TM \emph{site} carries three blocks of state, $u\in\R^{d_u}$, $s\in\R^{d_u}$ and $v\in\R^{d_v}$, with diagonal positive rates $\Lambda_u,\Lambda_s,\Lambda_v$ and gains $P=\diag(\rho)$. Between events,
\begin{equation}
\dot u=-\Lambda_u u,\qquad \dot s=\Lambda_s(Pu-s),\qquad \dot v=-\Lambda_v v.
\label{eq:drift}
\end{equation}
An event of type $c$ applies the jump
\begin{equation}
u\leftarrow u+\alpha_c,\qquad v\leftarrow v+g_c\,R_c\,s,
\label{eq:tmjump}
\end{equation}
with a positive scalar gate $g_c$ (in our implementation $g_c=1+\tanh(w_c^\top m_c)$, where $m_c$ are the event's intrinsic marks) and a read-out matrix $R_c\in\R^{d_v\times d_u}$. Events write $u$; the drift copies $u$ into $s$ with a delay; events read $s$ into $v$. No event writes $s$, and no event reads a block that events write, so all jumps commute (\cref{cor:affine}). The additive control (\ns) replaces the $v$-update by $v\leftarrow v+\beta_c$, and the selective control (\sel) by $v\leftarrow v+g_cR_cu$ evaluated before the $u$-update; these have the same state and drift. Over an interval of length $\Delta$ the drift~\eqref{eq:drift} has the closed form
\begin{equation}
\begin{gathered}
u\leftarrow e^{-\Lambda_u\Delta}u,\qquad v\leftarrow e^{-\Lambda_v\Delta}v,\\
s\leftarrow e^{-\Lambda_s\Delta}s+\Lambda_sP\,\varphi(\Lambda_s,\Lambda_u,\Delta)\,u,
\end{gathered}
\label{eq:closeddrift}
\end{equation}
with the elementwise function $\varphi(a,b,\Delta)=(e^{-a\Delta}-e^{-b\Delta})/(b-a)$, extended by continuity to $\Delta e^{-a\Delta}$ at $a=b$. A numerically stable evaluation is given in \cref{app:details}.

\subsection{Closed form and smoothness}

\begin{proposition}[Ordered-pair expansion]
\label{prop:closedform}
Let $K(\tau)=\Lambda_sP\,\varphi(\Lambda_s,\Lambda_u,\tau)$ for $\tau>0$ and $K(\tau)=0$ for $\tau\le0$. For any stream and query time $T$,
\begin{align*}
u(T)&=\textstyle\sum_i e^{-\Lambda_u(T-t_i)}\alpha_{c_i},\qquad s(T)=\sum_i K(T-t_i)\,\alpha_{c_i},\\
v(T)&=\textstyle\sum_j e^{-\Lambda_v(T-t_j)}\,g_{c_j}R_{c_j}\sum_i K(t_j-t_i)\,\alpha_{c_i}.
\end{align*}
Because $K$ vanishes for $\tau\le0$, the double sum in $v$ runs over ordered pairs $t_i<t_j$; events in the same tie group do not interact.
\end{proposition}

The accumulator $v$ is a learnable, multichannel version of the Hassenstein--Reichardt correlator~\cite{hassenstein1956}: a delayed, low-passed copy of one channel ($K*\alpha$) is multiplied by the arrival of another ($g_cR_c$). Equivalently, $v$ is an exponentially weighted iterated sum of order two over the counting paths of the event types, with the time letter forced between the two event letters.

\begin{lemma}
\label{lem:kernel}
For all $\tau$, $|K(\tau)|\le\rho$ and $|K'(\tau)|\le\lambda_s\rho$ elementwise; $K$ is continuous and Lipschitz on $\R$.
\end{lemma}

\begin{theorem}[Lipschitz continuity in the time stamps]
\label{thm:lipschitz}
Let $A=\max_c\|\alpha_c\|_\infty$, $\|R\|_\infty=\max_c\|R_c\|_\infty$ (row-sum norm), $g_c\le g_{\max}$, and let $\lambda_u,\lambda_s,\lambda_v,\rho$ denote the largest rates and gain. For a stream with $n$ events and any perturbation of the time stamps that keeps every event on the same side of the query time $T$, including perturbations that reorder events,
\begin{align*}
\|x(T;t)-x(T;t')\|_\infty&\le C\,\|t-t'\|_1,\\
C=\max\bigl\{\lambda_uA,\ \lambda_s\rho A,&\ g_{\max}\|R\|_\infty A\rho\,n(\lambda_v+2\lambda_s)\bigr\}.
\end{align*}
\end{theorem}

The constant grows with $n$ because each event can pair with up to $n$ others; for \ns the constant does not depend on $n$. An event that crosses the query time changes the set of included events, a discontinuity shared by every causal model. In the numerical checks of \cref{sec:experiments}, the observed change never exceeded $0.9\%$ of the bound.

\subsection{Separation from additive layers}

The target used throughout is the ordered-pair functional
\begin{equation}
\yord(S)=\sum_{i\in a,\ j\in b,\ t_i<t_j}\kappa(t_j-t_i),\qquad \kappa(\tau)=\tfrac{\tau}{20}\,e^{-\tau/20},
\label{eq:yord}
\end{equation}
with time in milliseconds, which sums a delay-tuned kernel over all pairs in which an $a$-event precedes a $b$-event.

\begin{theorem}[Additive layers cannot represent ordered pairs]
\label{thm:separation}
Consider any additive event model, that is, one with translation jumps $x\leftarrow x+b_c$ and linear drift, with finite state dimension $d$. Let its states be read at $q$ query times and passed to an arbitrary function $F$ (no continuity required). There is a constant $c>0$ such that for every $N\in\N$ there are two streams, each with at most $qd+2$ events of type $a$ and $N$ events of type $b$, on which $F$ returns the same value while $\yord$ differs by $Nc$. Hence $\sup_S|F(x(S))-\yord(S)|\ge Nc/2$, which is unbounded in $N$. The conclusion persists if the readout also sees, for each type, the time of its most recent event.
\end{theorem}

The proof (\cref{app:proofs}) fixes $N$ coincident $b$-events at a time $t^\ast$ that is not a query time. It then moves the $a$-events along a curve on which the additive state is exactly constant: an implicit-function argument applied to real-analytic impulse responses. Along this curve $\yord$ must change, because the derivative of $\kappa(t^\ast-\cdot)$ jumps at $t^\ast$ while no combination of analytic responses does. The statement covers additive layers of any width, any number of queries, and readouts with memory. It does not cover stacks of additive layers: a second layer whose marks depend on the first layer's state already forms ordered pairs (\cref{prop:height}).

\begin{proposition}[Three TM dimensions suffice]
\label{prop:tmyord}
Take $d_u=d_v=1$, $\lambda_s=\lambda_u=1/20$, $\rho=1$, $\alpha_a=1$, $\alpha_b=0$, $R_a=0$, $R_b=1$ and $g\equiv1$. Then $K=\kappa$ exactly and
\begin{gather*}
v(T)=\sum_{j\in b}e^{-\lambda_v(T-t_j)}\sum_{i\in a,\,t_i<t_j}\kappa(t_j-t_i),\\
0\le\yord-v(T)\le\bigl(1-e^{-\lambda_vW}\bigr)\yord,
\end{gather*}
where $W=T-t_0$ is the window length. As $\lambda_v\to0$, $v(T)\to\yord$; for $\lambda_v>0$, $v(T)$ is exactly the recency-weighted version of $\yord$.
\end{proposition}

\begin{remark}[Uniformity is what matters]
On a compact set of streams with a bounded number of events, additive features followed by a polynomial readout are already dense in the continuous tie-invariant functionals, by the Stone--Weierstrass theorem, because the per-type time moments separate such streams. \Cref{thm:separation} shows that this universality cannot be uniform in the number of events. The relevant comparison is therefore state dimension against event count, not universality.
\end{remark}

\subsection{Depth, higher-order layers and robust stacking}

We say a feature has \emph{nesting height} $h$ if it is built from $h$ nested causal sums of the form $\sum_{t_i\le t}w(t-t_i)f_i$; sums, products and pointwise functions of features take the maximum height of their arguments. In a stack, layer $\ell+1$ receives one event per tie group of the input, with a type determined by the input type and marks $z$ computed pointwise from the state of layer $\ell$.

\begin{proposition}[Two levels per TM layer]
\label{prop:height}
In a stack of $k$ additive layers every state coordinate has nesting height at most $k$; in a stack of $k$ TM layers, at most $2k$. The bound $2k$ is attained: with marks that are linear in the previous layer's $v$ and with $\lambda_v\to0$, $k$ TM layers represent the chain functional $\sum_{i_1<i_2\le i_3<i_4\le\cdots<i_{2k}}\prod_{m\ \mathrm{odd}}\kappa(t_{i_{m+1}}-t_{i_m})$ over a prescribed sequence of types.
\end{proposition}

\begin{conjecture}[Height separation]
\label{conj:height}
No architecture of nesting height $h$ can approximate a chain functional of height $h+1$ uniformly in the number of events. \Cref{thm:separation} is the case $h=1$.
\end{conjecture}

\Cref{prop:height} also corrects a natural expectation, that $k$ layers of order-aware updates reach order $2^k$. With nesting height as the measure, the provable rate is two levels per TM layer against one per additive layer. Counting summation indices after expanding products does not separate the two families either, because nonlinear inter-layer maps make that count grow exponentially for additive stacks as well.

Depth is not the only route to higher order. Order can also be stacked \emph{within} one layer, through the drift.

\begin{proposition}[TM-$k$: higher order in one layer]
\label{prop:tmk}
Extend the site to the chain $u\to s_1\Rightarrow v_1\to s_2\Rightarrow v_2\to\cdots\Rightarrow v_{k-1}$. Here $\to$ is drift ($\dot s_1=\Lambda_{s_1}(P_1u-s_1)$, $\dot s_{j+1}=\Lambda_{s_{j+1}}(P_{j+1}v_j-s_{j+1})$, with $u$ and each $v_j$ decaying), and $\Rightarrow$ is a jump ($u\leftarrow u+\alpha_c$, $v_j\leftarrow v_j+g_{c,j}R_{c,j}s_j$). All jumps commute, so the layer is partition invariant, tie invariant and crossing continuous. Its last block has the closed form
\[
v_{k-1}(T)=\sum_{i_1<i_2<\cdots<i_k}e^{-\Lambda_{v_{k-1}}(T-t_{i_k})}\,M_{k-1}\cdots M_1\,\alpha_{c_{i_1}},
\]
where $M_j=g_{c_{i_{j+1}},j}R_{c_{i_{j+1}},j}K_j(t_{i_{j+1}}-t_{i_j})$ and the kernels $K_j$ vanish for $\tau\le0$. One TM-$k$ layer thus represents ordered chains of $k$ events, of nesting height $k$, with state dimension linear in $k$.
\end{proposition}

The minimal TM layer is TM-2. Stacking is useful when the inter-layer map must be nonlinear, and the following rule keeps the guarantees.

\begin{proposition}[Robust stacking]
\label{prop:stacking}
Call a coordinate \emph{drift-mediated} if no jump writes it ($s$ and the $s_j$). In a TM-class layer, drift-mediated coordinates have no jumps at event times and are continuous functions of all time stamps, including across crossings. Hence, if layer $\ell+1$ receives events at the input event times, with types that depend only on the input types and with marks that are continuous functions of the drift-mediated coordinates of layer $\ell$, and if its own jumps commute for all values of the marks (as for TM layers), then the stack is partition invariant, tie invariant and crossing continuous. Marks read from event-written coordinates ($u$ or $v$) break crossing continuity, because they change by a full jump when another event joins or leaves the same tie group.
\end{proposition}

Information stored in $v_j$ is passed to the next layer through a drift-mediated copy $s_{j+1}$, as in TM-$k$; the chain construction of \cref{prop:height} then holds with one extra smoothing kernel per inter-layer link. A sparser alternative emits next-layer events only when a drift-mediated coordinate crosses a level (send-on-delta). This reduces the number of events but introduces thresholds, so \cref{cor:resolution} applies and continuity is lost at the levels.

\section{The Clock-Free Event (CFE) Layer}
\label{sec:cfe}

\begin{figure*}[t]
\centering
\includegraphics[width=\textwidth]{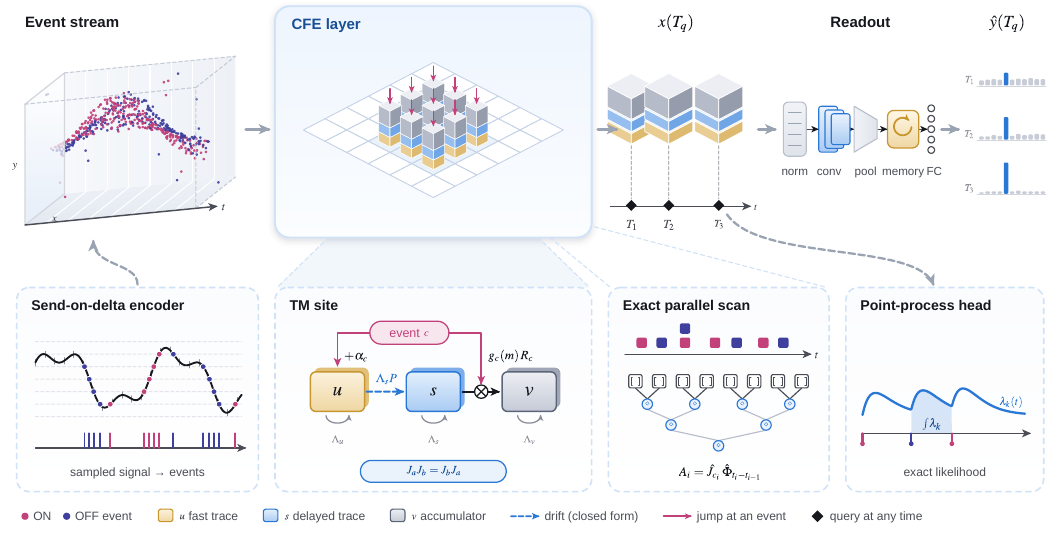}
\caption{The CFE network. An event stream (top left; ON and OFF events of a schematic rotating gesture), or a sampled signal converted into events by a send-on-delta encoder (bottom left, \cref{prop:sampling}), drives the CFE layer. The sensor is tiled into sites; an event (red pixel) applies a jump to the state of every site in its $3\times3$ neighbourhood, with a letter given by its polarity and relative offset, and every site state drifts in closed form between events. \emph{TM site:} events write the fast trace $u$ and, through the mark-dependent gate $g_c(m)R_c$, read the delayed trace $s$ into the accumulator $v$; no event writes a coordinate that events read, so all jumps commute (\cref{cor:affine}). \emph{Exact parallel scan:} each event acts as an affine map, so the states are computed exactly by a segmented parallel scan, and tie groups may be processed in any order. Queries read the site states $x(T_q)$ at arbitrary times; a readout shared by all arms (signed-log normalization with raw features, convolution over the site grid, anchor-based pooling, a memory across queries and a linear layer) gives a prediction at every query. For event sequences without topology, an intensity head with a closed-form compensator gives the exact likelihood (\cref{prop:compensator}). Bar heights are schematic.}
\label{fig:cfe}
\end{figure*}

\subsection{Sites, letters and marks}

\Cref{fig:cfe} shows the network end to end. For sensors with spatial or spectral topology, the CFE layer tiles the sensor into \emph{sites}. For a $128\times128$ event camera we use $16\times16$ sites of $8\times8$ pixels (configuration V1) or $32\times32$ sites of $4\times4$ pixels (V2); for the 700 cochlear channels of SHD we use sites of 8 adjacent channels. An event acts on every site in the $3\times3$ (respectively 3-site) neighbourhood of its own site. Its \emph{letter} at a site is the pair (polarity, relative offset), which gives 18 letters on the camera; on SHD the polarity bit is replaced by the parity of the channel within its site, giving 6 letters. Letter parameters $(\alpha,R,g,\beta)$ are shared across sites, which makes the layer translation equivariant in the manner of a convolution. Each site holds a TM state as in \cref{sec:tm}, with $d_u=d_s=8$ and $d_v=16$, and its own last-update time.

Marks are restricted to quantities intrinsic to the event, here the position of the pixel or channel within its site. Marks computed from the history of other events, such as the time since the last event at a pixel, would make the jump maps depend on the processing history and would reintroduce order and tie dependence.

\paragraph{Topology-free variant.} For event sequences without spatial structure, such as marked temporal point processes, irregularly sampled multivariate time series or multi-electrode spike trains, the layer uses one local site per event type and one global site that receives every event. The letter is then the event type, and observed values enter as marks through a fixed embedding.

\subsection{Exact computation}

Time stamps are kept as integers (microseconds for the sensors used here), and events with equal stamps form a tie group. A query at time $T$ includes all events with $t_i\le T$. Each analysis window starts from a cold state, and no event before the window start enters it. The drift~\eqref{eq:closeddrift} is evaluated in closed form. With $z=(b-a)\Delta$, $\varphi(a,b,\Delta)=\Delta e^{-a\Delta}E(z)$ for $z\ge0$, with the mirror form for $z<0$, and $E(z)=-\mathrm{expm1}(-z)/z$ (Taylor series for $|z|<10^{-3}$). This expression is accurate for nearly equal rates and for long gaps.

For training, the per-site composition of drift and jump over an event is an affine map, and composition of affine maps is associative. The state sequence can therefore be computed with a segmented parallel scan, with segments reset at window starts, and backpropagated with the adjoint scan. Chunk sizes affect only speed; we verify agreement with an FP64 event-by-event reference, with chunk boundaries chosen at random. Queries are inserted as read-only pseudo-events. A query drifts a copy of each site's state to $T$ and does not modify it, so queries can be placed at arbitrary times: the layer itself has no time step.

\subsection{Readout and anytime training}

At each query time, the site states $[u,s,v]$ are compressed by a signed logarithm, normalized block-wise by their root mean square with learnable gains, and concatenated with eight raw per-site query features: for each polarity, pixel recency at three time constants and a log-compressed decayed count. A convolutional head over the site grid, followed by anchor-based spatial pooling, a bounded memory across query times and a linear classifier, produces the logits. All arms share this readout; the readouts for SHD and Speech Commands are analogous (\cref{app:details}). Because the memory integrates over query times, the network output depends on the query schedule (here every 50\,ms); it does not depend on how events are partitioned.

We train for \emph{anytime} classification: the loss averages the cross-entropy over a set of prefixes (13 prefixes from 200 to 800\,ms on DVS128 Gesture, 10 on SHD). A single model then gives usable predictions at every prefix. For fairness, we train clocked baselines both with their original recipe and with the same multi-prefix objective, and report the better of the two at each prefix (\cref{sec:protocol}).

\subsection{Cost}

Per input event and per affected site, a TM update costs $O(d_vd_u+d_ud_m)$ multiply--accumulates and $O(d_u+d_s+d_v)$ exponential-type evaluations, where $d_m$ is the number of intrinsic mark dimensions; the cost is independent of the sensor resolution and of any clock. With $d_u=d_s=8$, $d_v=16$, two mark dimensions and at most nine sites per event, this is about $1.9\times10^3$ multiply--accumulates and $3.7\times10^2$ exponential-type evaluations per input event for TM, and about $0.9\times10^3$ multiply--accumulates for NS. These counts cover the layer only; the readout runs only at query times, and its cost is amortized over the events between queries. The persistent state is $d_u+d_s+d_v=32$ numbers and one time stamp per site.

\subsection{From sampled signals to events}
\label{sec:encoder}

Signals that arrive on a sampling grid, such as audio or inertial measurements, can be converted into events by a send-on-delta (level-crossing) encoder. For each channel, an event of sign $\pm1$ is emitted whenever the linearly interpolated signal crosses one of the levels $r_0+m\theta$, and it is time-stamped at the interpolated crossing time. The events approximate those of the underlying continuous signal, so the sampling rate becomes a detail of acquisition rather than a parameter of the model.

\begin{proposition}[Sampling-rate consistency]
\label{prop:sampling}
Let $x\in C^2[0,W]$ with $\|x''\|_\infty\le M$, and suppose every level crossing of $x$ is transversal, with $|x'|\ge\mu>0$ at each crossing. For all sufficiently small sampling intervals $h$, the encoder applied to the samples $x(kh)$ emits the same sequence of (level, sign) events as the continuous signal, and each time stamp is off by at most $Mh^2/\bigl(8(\mu-Mh)\bigr)$. Consequently, a TM layer fed with these events has state error at most $C\,n\,Mh^2/\bigl(8(\mu-Mh)\bigr)$, with $C$ as in \cref{thm:lipschitz} and $n$ the number of events.
\end{proposition}

Training at one sampling rate and deploying at another therefore changes the input of a CFE network by $O(h^2)$ in its time stamps. A frame-based network, by contrast, sees a different tensor at every rate. Sampling-frequency-independent convolutions~\cite{saito2021sfi} address the same problem for audio by other means.

\subsection{Temporal point processes with closed-form likelihood}

In the topology-free variant, an intensity head turns the state into conditional intensities $\lambda_k(t)$ for each event type. The log-likelihood requires the compensator $\int\lambda_k\,dt$ between events, which is usually approximated by quadrature or Monte Carlo sampling. In a development run on a public e-commerce sequence dataset, we found that 20-node quadrature shifted the per-event log-likelihood of our models by up to 0.065 nats relative to an adaptive FP64 integral.

\begin{proposition}[Exact compensator]
\label{prop:compensator}
Let $\lambda_k(t)=\mu_k+h(t)^\top Q_kh(t)$ with $\mu_k\ge0$ and $Q_k\succeq0$, where $h(t)$ is a fixed linear function of the TM state. Between consecutive events, every entry of $h(t)h(t)^\top$ is a finite combination of functions $\tau^pe^{-\gamma\tau}$ with $p\le2$ and $\gamma\ge0$. Hence $\lambda_k\ge0$, and $\int_{t_n}^{t}\lambda_k$ and the exact log-likelihood are available in closed form.
\end{proposition}

\section{A Preregistered Protocol for Clock Robustness}
\label{sec:protocol}

The claims of \cref{sec:theory,sec:tm} make specific empirical predictions. We fixed the following protocol before evaluating on any official test set. Its gates and hypotheses were recorded with SHA-256 digests before any test evaluation, and each evaluation script is frozen in the same way before its test set is read.

\paragraph{Single-use test sets.} Model development uses only official training data. For DVS128 Gesture we use four subject-disjoint folds of the official training split, each with a calibration subset (99, 143, 99 and 154 clips, 495 in total). For SHD we hold out about 10\% of the training speakers as a validation set. Each official test set is read once, by a frozen evaluation script, after the models, the checkpoint list and the analysis have been frozen. Shared random perturbations are generated by a frozen, deterministic, order-independent generator. Disclosure: before this study, an earlier and unrelated generation of our models was evaluated once on the 263 eligible clips of the DVS128 Gesture test split; no model or decision in this paper uses that evaluation.

\paragraph{Preregistered hypotheses (DVS128 Gesture).} At the 200\,ms prefix, compared by paired bootstrap ($10^4$ resamples): the TM model is more accurate than the additive model (H-G1), than the clocked spiking baseline PLIF~\cite{fang2021plif} (H-G2), and than Event-SSM~\cite{schoene2024eventssm} (H-G3). Under shared time-stamp jitter ($\sigma=1$ and $5$\,ms), the accuracy drop of TM is smaller than that of either PLIF variant (H-R1). We also report all 13 prefixes and complete-clip predictions, obtained by averaging the logits of all queries from 200\,ms to the last query.

\paragraph{Fair baselines.} Each baseline is trained (i) with its original recipe and (ii) with our multi-prefix objective, and the better of the two is used at each prefix. Baselines use their official code where available, with the same data splits and selection rules as our models. Compute-matched baselines are obtained by shrinking channel widths until the dense multiply--accumulate count matches ours. Compute is reported in three ways: dense, sparse (synaptic operations) and measured.

\paragraph{Clock-shift conditions.} Following the frequency-generalization protocol of~\cite{zubic2024ssm}, a trained model is evaluated without retraining under changes that alter the clock or the slicing but not the signal:
\begin{itemize}
\item frame count halved and doubled, and the frame grid shifted by half a frame (clocked models);
\item packetization into single events, 1024-event chunks and 10\,ms chunks (event models);
\item a change of sampling rate, 8\,kHz and 44.1\,kHz for models trained at 16\,kHz, with and without resampling back to the training rate (audio);
\item shuffling within tie groups;
\item Gaussian time-stamp jitter;
\item random deletion of 10\%, 30\% and 50\% of the events.
\end{itemize}
The reported quantity is the drop in accuracy relative to the clean condition. For CFE, the packetization conditions test an invariance that holds by construction, so they serve as a check of the implementation; the other conditions are genuine tests.

\paragraph{Retiming attacks in physical time.} Spike-retiming attacks~\cite{yu2026retiming} are usually budgeted in time bins. \Cref{prop:binned} shows that a bin-level budget conceals the real cost: in physical time, an arbitrarily small displacement suffices to move an event across a bin edge. We therefore specify budgets in physical time. Under $B_\infty(\varepsilon)$, every time stamp moves by at most $\varepsilon\in\{0.1,1,5\}$\,ms; under $B_0(k,\varepsilon)$, at most a fraction $k\in\{1\%,2.5\%\}$ of the events move, each by at most $\varepsilon$. Counts, locations and polarities are unchanged. Event models are attacked by projected gradient descent on the time stamps and by a gradient-free random search, the latter to rule out gradient masking. Binned models are attacked by moving the events nearest to bin edges. The theory predicts that attack success against a TM model scales with the total displacement (\cref{thm:lipschitz}), whereas against a binned model it is governed by the fraction of events within $\varepsilon$ of an edge.

\section{Preliminary Experiments}
\label{sec:experiments}

This section reports numerical checks of the theory and development-set results. All accuracies below come from development folds or validation splits, and model selection used the same data, so absolute numbers are optimistic. Comparisons between arms are nonetheless like-for-like: every arm uses the same readout, recipe, selection rule and data. Official test results will follow the protocol of \cref{sec:protocol}.

\subsection{Numerical verification of the theory}

\begin{figure}[t]
\centering
\includegraphics[width=\columnwidth]{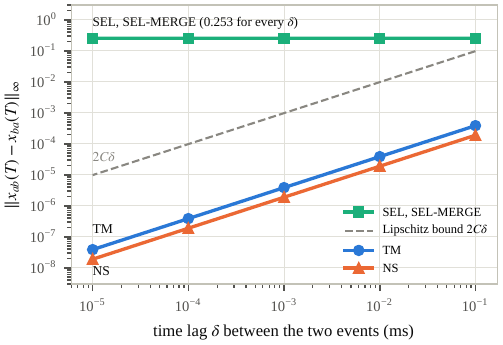}
\caption{State difference when two events of different types swap places at lag $\delta$ (FP64). TM and NS change linearly in $\delta$, below the Lipschitz bound of \cref{thm:lipschitz}. Selective updates jump by a constant amount at every lag, as \cref{cor:norescue} predicts.}
\label{fig:swap}
\end{figure}

We implemented a single TM site independently of the training code, in FP64, with $d_u=d_s=3$, $d_v=2$, two event types and randomly drawn parameters (\cref{app:details}). \Cref{tab:checks} summarizes six checks. \Cref{fig:swap} shows the central contrast. When two events of different types swap places at lag $\delta$, the TM and NS states change linearly in $\delta$, within the Lipschitz bound, whereas the selective updates (\sel and its group-update variant \selm) jump by a fixed amount however small $\delta$ is. At an exact tie, \selm returns the same state for both orders and \sel does not.

\begin{table}[t]
\centering
\caption{Numerical checks of the theory (FP64, seed 23).}
\label{tab:checks}
\small
\setlength{\tabcolsep}{3pt}
\begin{tabular}{@{}>{\raggedright\arraybackslash}p{0.30\columnwidth}>{\raggedright\arraybackslash}p{0.42\columnwidth}>{\raggedright\arraybackslash}p{0.2\columnwidth}@{}}
\toprule
Check & Result & Supports \\
\midrule
Closed form vs.\ recursion (200 streams, 5--60 events, $\approx$30\% ties) & max.\ relative error $2.5\times10^{-15}$ & Prop.~\ref{prop:closedform} \\
Shuffling within ties (100 streams, $\approx$60\% ties) & max.\ $|\Delta x|$: TM $1.7\times10^{-13}$, NS $1.4\times10^{-14}$, \selm $3.4\times10^{-13}$, \sel $43.5$ & Thm.~\ref{thm:commute} \\
Time-stamp perturbations $10^{-4}$--$3.2$\,ms, with reordering (300) & observed/bound $\le0.0089$ & Thm.~\ref{thm:lipschitz} \\
Swap at lag $\delta\in[10^{-5},10^{-1}]$\,ms & TM $3.84\times10^{-3}\delta$; NS $1.88\times10^{-3}\delta$; \sel, \selm 0.2526--0.2528 & Cor.~\ref{cor:norescue}, Thm.~\ref{thm:timemediated} \\
NS fiber ($d=5$; 50 $b$-events) & state residual $\le6.7\times10^{-16}$ while $\yord$ falls from 37.00 to 32.30 & Thm.~\ref{thm:separation} \\
TM represents $\yord$ (200 streams) & $|K-\kappa|\le1.1\times10^{-16}$; all errors within bound; max.\ 6.4\% (bound 9.5\%) & Prop.~\ref{prop:tmyord} \\
\bottomrule
\end{tabular}
\end{table}

\subsection{DVS128 Gesture (development folds)}

\begin{figure}[t]
\centering
\includegraphics[width=\columnwidth]{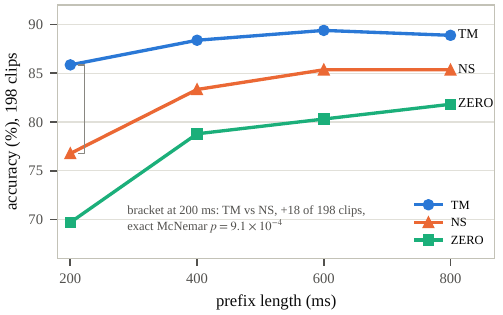}
\caption{DVS128 Gesture, folds 0 and 2 (198 calibration clips): accuracy against observation window for TM, NS and the stateless arm (\zero). The benefit of the ordered state is largest early.}
\label{fig:dvs}
\end{figure}

DVS128 Gesture~\cite{amir2017gesture} contains 11 hand-gesture classes recorded with a $128\times128$ event camera. We use 800\,ms windows with cold start, time stamps at 1\,\textmu s resolution, queries every 50\,ms and the 13-prefix objective. Arms differ only in the site update (\cref{sec:tm}); the readout, training recipe and selection rule (lowest calibration cross-entropy at 800\,ms) are shared. \Cref{tab:dvs} and \cref{fig:dvs} summarize the results.

On two folds (198 clips), TM exceeds NS by 18 clips at 200\,ms (exact McNemar test, 23 versus 5 discordant clips, $p=9.1\times10^{-4}$, significant after Bonferroni correction over eight comparisons). The margins at 400, 600 and 800\,ms are 10, 8 and 7 clips ($p=0.052$, $0.077$ and $0.143$). TM exceeds the stateless arm at all four prefixes ($p$ between $9.4\times10^{-7}$ and $0.0043$, all significant after correction). The advantage of order is largest early: the order of events carries information within a short time, and unordered statistics need longer to catch up. On all four folds (495 clips), TM exceeds the better of NS and \zero at each prefix by 40, 12, 22 and 21 clips. On the trained models, packetization changed no prediction, and shuffling ties changed no prediction for TM, NS and \zero.

On fold 0, a PLIF network trained with the multi-prefix objective is level with TM (77/80/85/85 versus 79/82/86/84 correct out of 99). The same architecture trained with its original full-clip recipe reaches 83/99 on complete clips but only 17/99 at 200\,ms and 32/99 at 800\,ms, so clocked models need prefix-aware training to be usable at arbitrary times. With finer sites and augmentation (V2), TM reaches 170/177/179/179 against 162/174/174/175 for NS.

\begin{table}[t]
\centering
\caption{DVS128 Gesture, development folds: correct clips at the 200/400/600/800\,ms prefixes. V1: $8\times8$-pixel sites; V2: $4\times4$-pixel sites with augmentation. ``Best unordered'' is the better of NS and \zero at each prefix.}
\label{tab:dvs}
\small
\setlength{\tabcolsep}{3.5pt}
\begin{tabular}{@{}llcccc@{}}
\toprule
Setting (clips) & Arm & 200 & 400 & 600 & 800 \\
\midrule
\multirow{3}{*}{V1, folds 0, 2 (198)} & TM & \textbf{170} & \textbf{175} & \textbf{177} & \textbf{176} \\
 & NS & 152 & 165 & 169 & 169 \\
 & \zero & 138 & 156 & 159 & 162 \\
\midrule
\multirow{2}{*}{V1, folds 0--3 (495)} & TM & \textbf{407} & \textbf{419} & \textbf{440} & \textbf{440} \\
 & best unordered & 367 & 407 & 418 & 419 \\
\midrule
\multirow{2}{*}{V2, folds 0, 2 (198)} & TM & \textbf{170} & \textbf{177} & \textbf{179} & \textbf{179} \\
 & NS & 162 & 174 & 174 & 175 \\
\midrule
\multirow{5}{*}{fold 0 (99)} & TM (V1) & \textbf{79} & \textbf{82} & \textbf{86} & 84 \\
 & NS (V1) & 67 & 77 & 80 & 83 \\
 & \zero & 61 & 72 & 75 & 76 \\
 & PLIF, multi-prefix & 77 & 80 & 85 & \textbf{85} \\
 & PLIF, original recipe & 17 & -- & -- & 32 \\
\bottomrule
\end{tabular}
\end{table}

\subsection{Spiking Heidelberg Digits and Speech Commands}

SHD~\cite{cramer2022shd} contains spoken digits (20 classes) encoded as spikes on 700 cochlear channels. We evaluate on a speaker-held-out validation split of the official training set, at 10 query times. Speech Commands~\cite{warden2018speech} (35 classes) is converted into channel-wise events by a fixed encoder that is shared by all arms. \Cref{tab:audio} reports the results. On both datasets the ordered TM state is far ahead of the additive state: by 24.5 points on SHD and by 30.6 points on Speech Commands. On Speech Commands the TM model is still training and trails a frame-based depthwise-separable CNN (DS-CNN)~\cite{zhang2017helloedge} trained on the same split by 9.0 points. On SHD our validation accuracy is well below published test accuracies of 95--96\%~\cite{hammouamri2024dcls,schoene2024eventssm}; those were obtained under different protocols and are not directly comparable. In a site-size study on SHD (\cref{app:shd}), accuracy peaks at 8 channels per site, and a fourfold wider state did not help.

\begin{table}[t]
\centering
\caption{Audio benchmarks, validation splits (accuracy, \%). SHD: mean over 10 query times and at 1000\,ms. Speech Commands (SC): 9,981 validation clips; TM and NS are still training (epoch 20).}
\label{tab:audio}
\small
\setlength{\tabcolsep}{4pt}
\begin{tabular}{@{}llcc@{}}
\toprule
Dataset, configuration & Arm & 10-point mean & final \\
\midrule
\multirow{3}{*}{SHD, augmented (A)} & TM & \textbf{81.80} & \textbf{87.40} \\
 & NS & 64.31 & 62.88 \\
 & \zero & -- & 8.73 \\
\midrule
\multirow{3}{*}{SHD, base} & TM & \textbf{73.25} & \textbf{76.45} \\
 & NS & 51.59 & 47.37 \\
 & \zero & 11.36 & 8.86 \\
\midrule
\multirow{3}{*}{SC (35 classes)} & TM & -- & 79.56 \\
 & NS & -- & 48.98 \\
 & DS-CNN & -- & \textbf{88.55} \\
\bottomrule
\end{tabular}
\end{table}

\subsection{An observation on clocked models}

Re-evaluating the same PLIF checkpoint on two GPU platforms (NVIDIA A800 and V100), with identical inputs, batch size and deterministic settings, changed the prediction for 11 of 99 calibration clips at one or more prefixes (27 prefix predictions in total; largest logit difference 0.18). Because the inputs, the batch size and the evaluation code were identical, the most likely cause is that spiking thresholds amplify differences in floating-point arithmetic, which is the discontinuity that \cref{cor:resolution} attributes to such models. The corresponding cross-platform comparison for CFE is part of the protocol.

\section{Discussion and Limitations}
\label{sec:discussion}

\paragraph{What the theory does and does not say.} \Cref{thm:commute} is an equivalence that holds for all event models with continuous drift and jumps. It explains why gated and selective event updates cannot be made robust to ties and jitter by any tie rule or augmentation, short of making them commute. \Cref{thm:timemediated} is a first-order statement: it identifies where order information can come from once updates commute and how it scales at small lags. It does not bound the amount of order information at larger lags, which depends on the time constants. The Lipschitz constant of \cref{thm:lipschitz} grows with the number of events, so for long dense streams the guarantee is qualitative rather than tight. \Cref{thm:separation} concerns single additive layers; whether its height generalization (\cref{conj:height}) holds is open.

\paragraph{Relation to signature models.} Interval composition, and hence partition invariance, is shared with online log-signature models~\cite{walker2026faithful}, and exponentially weighted iterated sums are studied in~\cite{bloch2026ews}. Our contribution is the characterization of what robustness to ties and crossings requires, the time-mediated structure that remains order-aware under that requirement, and a local, weight-shared instantiation whose cost is independent of the number of sensor channels. In contrast, log-signature tensors grow as $d^N$ with the number of channels $d$ and truncation level $N$. Whether the TM class is universal, in a quantitative sense, for continuous tie-invariant functionals, for instance through a partially commutative signature in which event letters commute with each other but not with the time letter, is an open question.

\paragraph{Accuracy.} Our development results do not establish state-of-the-art accuracy. On SHD we are well below published test accuracies, and on Speech Commands below a compact CNN. On DVS128 Gesture, a clocked spiking baseline trained with our multi-prefix objective is level with TM on one fold. A second layer did not yet improve over a single layer in our DVS experiments, which were stopped early, so in this version the depth results of \cref{prop:height,prop:tmk} are supported by theory only. The protocol of \cref{sec:protocol} is designed to measure what the theory predicts: accuracy at equal compute, behavior under changes of clock and slicing, and robustness to retiming.

\paragraph{Readout and query schedule.} The CFE layer has no time step, but a readout with memory across query times makes predictions depend on the query schedule. Readouts that are themselves continuous-time, for example a TM layer whose queries only read state, would remove this dependence.

\section{Conclusion}

Partition invariance, independence from how a stream is sliced, follows from exact event-driven computation. Robustness to simultaneous events and to timing jitter requires more: the event updates must commute, and then all sensitivity to order must be mediated by the dynamics between events. This yields a precise design principle, a minimal layer that realizes it with ordered pair interactions, constructions for higher order and depth that preserve it, and an evaluation protocol that tests it. We hope that time-mediated non-commutativity will be useful as a design constraint for learning systems that consume asynchronous signals directly, without a clock.

{\small
\bibliographystyle{unsrtnat}
\bibliography{refs}

\begin{thebibliography}{43}
\providecommand{\natexlab}[1]{#1}
\providecommand{\url}[1]{\texttt{#1}}
\expandafter\ifx\csname urlstyle\endcsname\relax
  \providecommand{\doi}[1]{doi: #1}\else
  \providecommand{\doi}{doi: \begingroup \urlstyle{rm}\Url}\fi

\bibitem[Lichtsteiner et~al.(2008)Lichtsteiner, Posch, and
  Delbruck]{lichtsteiner2008}
Patrick Lichtsteiner, Christoph Posch, and Tobi Delbruck.
\newblock A $128\times128$ 120 {dB} 15 $\mu$s latency asynchronous temporal
  contrast vision sensor.
\newblock \emph{IEEE Journal of Solid-State Circuits}, 43\penalty0
  (2):\penalty0 566--576, 2008.

\bibitem[Gallego et~al.(2022)Gallego, Delbr{\"u}ck, Orchard, Bartolozzi, Taba,
  Censi, Leutenegger, Davison, Conradt, Daniilidis, and
  Scaramuzza]{gallego2022survey}
Guillermo Gallego, Tobi Delbr{\"u}ck, Garrick Orchard, Chiara Bartolozzi, Brian
  Taba, Andrea Censi, Stefan Leutenegger, Andrew~J. Davison, J{\"o}rg Conradt,
  Kostas Daniilidis, and Davide Scaramuzza.
\newblock Event-based vision: A survey.
\newblock \emph{IEEE Transactions on Pattern Analysis and Machine
  Intelligence}, 44\penalty0 (1):\penalty0 154--180, 2022.

\bibitem[Cramer et~al.(2022)Cramer, Stradmann, Schemmel, and
  Zenke]{cramer2022shd}
Benjamin Cramer, Yannik Stradmann, Johannes Schemmel, and Friedemann Zenke.
\newblock The {H}eidelberg spiking data sets for the systematic evaluation of
  spiking neural networks.
\newblock \emph{IEEE Transactions on Neural Networks and Learning Systems},
  33\penalty0 (7):\penalty0 2744--2757, 2022.

\bibitem[Fang et~al.(2021)Fang, Yu, Chen, Masquelier, Huang, and
  Tian]{fang2021plif}
Wei Fang, Zhaofei Yu, Yanqi Chen, Timoth{\'e}e Masquelier, Tiejun Huang, and
  Yonghong Tian.
\newblock Incorporating learnable membrane time constant to enhance learning of
  spiking neural networks.
\newblock In \emph{Proceedings of the IEEE/CVF International Conference on
  Computer Vision (ICCV)}, 2021.

\bibitem[Amir et~al.(2017)]{amir2017gesture}
Arnon Amir et~al.
\newblock A low power, fully event-based gesture recognition system.
\newblock In \emph{Proceedings of the IEEE Conference on Computer Vision and
  Pattern Recognition (CVPR)}, 2017.

\bibitem[Chowdhury et~al.(2026)Chowdhury, Sharmin, and
  Ahmed]{chowdhury2026tradeoff}
Sayeed~Shafayet Chowdhury, Ruhi Sharmin, and Syed~Ishtiaque Ahmed.
\newblock On the robustness-resolution tradeoff in temporal quantization of
  event streams.
\newblock \emph{arXiv preprint arXiv:2609.22295}, 2026.

\bibitem[Zubi{\'c} et~al.(2024)Zubi{\'c}, Gehrig, and Scaramuzza]{zubic2024ssm}
Nikola Zubi{\'c}, Mathias Gehrig, and Davide Scaramuzza.
\newblock State space models for event cameras.
\newblock In \emph{Proceedings of the IEEE/CVF Conference on Computer Vision
  and Pattern Recognition (CVPR)}, 2024.

\bibitem[Yu et~al.(2026)Yu, Zhang, Ye, Lin, Wei, Wang, Yang, Tao, and
  Jiang]{yu2026retiming}
Yi~Yu, Qixin Zhang, Shuhan Ye, Xun Lin, Qianshan Wei, Kun Wang, Wenhan Yang,
  Dacheng Tao, and Xudong Jiang.
\newblock Time is all it takes: Spike-retiming attacks on event-driven spiking
  neural networks.
\newblock In \emph{International Conference on Learning Representations
  (ICLR)}, 2026.

\bibitem[Messikommer et~al.(2020)Messikommer, Gehrig, Loquercio, and
  Scaramuzza]{messikommer2020asynet}
Nico Messikommer, Daniel Gehrig, Antonio Loquercio, and Davide Scaramuzza.
\newblock Event-based asynchronous sparse convolutional networks.
\newblock In \emph{Proceedings of the European Conference on Computer Vision
  (ECCV)}, 2020.

\bibitem[Schaefer et~al.(2022)Schaefer, Gehrig, and
  Scaramuzza]{schaefer2022aegnn}
Simon Schaefer, Daniel Gehrig, and Davide Scaramuzza.
\newblock {AEGNN}: Asynchronous event-based graph neural networks.
\newblock In \emph{Proceedings of the IEEE/CVF Conference on Computer Vision
  and Pattern Recognition (CVPR)}, 2022.

\bibitem[Sekikawa et~al.(2019)Sekikawa, Hara, and Saito]{sekikawa2019eventnet}
Yusuke Sekikawa, Kosuke Hara, and Hideo Saito.
\newblock {EventNet}: Asynchronous recursive event processing.
\newblock In \emph{Proceedings of the IEEE/CVF Conference on Computer Vision
  and Pattern Recognition (CVPR)}, 2019.

\bibitem[Sch{\"o}ne et~al.(2024)Sch{\"o}ne, Sushma, Zhuge, Mayr, Subramoney,
  and Kappel]{schoene2024eventssm}
Mark Sch{\"o}ne, Neeraj~Mohan Sushma, Jingyue Zhuge, Christian Mayr, Anand
  Subramoney, and David Kappel.
\newblock Scalable event-by-event processing of neuromorphic sensory signals
  with deep state-space models.
\newblock \emph{arXiv preprint arXiv:2404.18508}, 2024.

\bibitem[Hao et~al.(2026)Hao, Sui, Zou, Dai, Zubi{\'c}, Scaramuzza, and
  Wang]{hao2026ssla}
Haiqing Hao, Zhipeng Sui, Rong Zou, Zijia Dai, Nikola Zubi{\'c}, Davide
  Scaramuzza, and Wenhui Wang.
\newblock Low-latency event-based object detection with spatially-sparse linear
  attention.
\newblock \emph{arXiv preprint arXiv:2603.06228}, 2026.

\bibitem[Pellerito et~al.(2026)Pellerito, Gehrig, Shiba, and
  Scaramuzza]{pellerito2026neuralevents}
Roberto Pellerito, Daniel Gehrig, Shintaro Shiba, and Davide Scaramuzza.
\newblock Neural events: Discrete asynchronous autoencoders for event-based
  vision.
\newblock \emph{arXiv preprint arXiv:2606.19835}, 2026.

\bibitem[Walker et~al.(2026)Walker, Bloch, Yang, Morley, and
  Lyons]{walker2026faithful}
Benjamin Walker, Alexandre Bloch, Lingyi Yang, Sam Morley, and Terry Lyons.
\newblock Faithful embeddings of irregular and asynchronous data for online
  {Log-NCDEs}.
\newblock \emph{arXiv preprint arXiv:2605.30213}, 2026.

\bibitem[Bloch et~al.(2026)Bloch, Walker, Mouterde, Morley, Cohen, and
  Lyons]{bloch2026ews}
Alexandre Bloch, Benjamin Walker, Jo{\"e}l Mouterde, Sam Morley, Samuel~N.
  Cohen, and Terry Lyons.
\newblock Extending {SSM}s with the exponentially weighted signature.
\newblock \emph{arXiv preprint arXiv:2603.19198}, 2026.

\bibitem[Hassenstein and Reichardt(1956)]{hassenstein1956}
Bernhard Hassenstein and Werner Reichardt.
\newblock Systemtheoretische {A}nalyse der {Z}eit-, {R}eihenfolgen- und
  {V}orzeichenauswertung bei der {B}ewegungsperzeption des
  {R}{\"u}sselk{\"a}fers {C}hlorophanus.
\newblock \emph{Zeitschrift f{\"u}r Naturforschung B}, 11\penalty0
  (9--10):\penalty0 513--524, 1956.

\bibitem[Warden(2018)]{warden2018speech}
Pete Warden.
\newblock Speech commands: A dataset for limited-vocabulary speech recognition.
\newblock \emph{arXiv preprint arXiv:1804.03209}, 2018.

\bibitem[Fang and Panda(2026)]{fang2026event2vec}
Wei Fang and Priyadarshini Panda.
\newblock {Event2Vec}: Processing neuromorphic events directly by
  representations in vector space.
\newblock In \emph{Proceedings of the 43rd International Conference on Machine
  Learning (ICML)}, 2026.

\bibitem[Fang et~al.(2023)]{fang2023spikingjelly}
Wei Fang et~al.
\newblock {SpikingJelly}: An open-source machine learning infrastructure
  platform for spike-based intelligence.
\newblock \emph{Science Advances}, 9\penalty0 (40), 2023.

\bibitem[Wunderlich and Pehle(2021)]{wunderlich2021eventprop}
Timo~C. Wunderlich and Christian Pehle.
\newblock Event-based backpropagation can compute exact gradients for spiking
  neural networks.
\newblock \emph{Scientific Reports}, 11:\penalty0 12829, 2021.

\bibitem[Hammouamri et~al.(2024)Hammouamri, Khalfaoui-Hassani, and
  Masquelier]{hammouamri2024dcls}
Ilyass Hammouamri, Ismail Khalfaoui-Hassani, and Timoth{\'e}e Masquelier.
\newblock Learning delays in spiking neural networks using dilated convolutions
  with learnable spacings.
\newblock In \emph{International Conference on Learning Representations
  (ICLR)}, 2024.

\bibitem[Morrill et~al.(2026)Morrill, Pehle, and Zador]{morrill2026bullet}
Todd Morrill, Christian Pehle, and Anthony Zador.
\newblock Bullet trains: Parallelizing training of temporally precise spiking
  neural networks.
\newblock \emph{arXiv preprint arXiv:2603.13283}, 2026.

\bibitem[Subramoney et~al.(2023)Subramoney, Nazeer, Sch{\"o}ne, Mayr, and
  Kappel]{subramoney2023egru}
Anand Subramoney, Khaleelulla~Khan Nazeer, Mark Sch{\"o}ne, Christian Mayr, and
  David Kappel.
\newblock Efficient recurrent architectures through activity sparsity and
  sparse back-propagation through time.
\newblock In \emph{International Conference on Learning Representations
  (ICLR)}, 2023.

\bibitem[Rubanova et~al.(2019)Rubanova, Chen, and
  Duvenaud]{rubanova2019latentode}
Yulia Rubanova, Ricky T.~Q. Chen, and David Duvenaud.
\newblock Latent ordinary differential equations for irregularly-sampled time
  series.
\newblock In \emph{Advances in Neural Information Processing Systems
  (NeurIPS)}, 2019.

\bibitem[De~Brouwer et~al.(2019)De~Brouwer, Simm, Arany, and
  Moreau]{debrouwer2019gruodebayes}
Edward De~Brouwer, Jaak Simm, Adam Arany, and Yves Moreau.
\newblock {GRU-ODE-Bayes}: Continuous modeling of sporadically-observed time
  series.
\newblock In \emph{Advances in Neural Information Processing Systems
  (NeurIPS)}, 2019.

\bibitem[Kidger et~al.(2020)Kidger, Morrill, Foster, and Lyons]{kidger2020ncde}
Patrick Kidger, James Morrill, James Foster, and Terry Lyons.
\newblock Neural controlled differential equations for irregular time series.
\newblock In \emph{Advances in Neural Information Processing Systems
  (NeurIPS)}, 2020.

\bibitem[Walker et~al.(2024)]{walker2024logncde}
Benjamin Walker et~al.
\newblock Log neural controlled differential equations: The {L}ie brackets make
  a difference.
\newblock In \emph{Proceedings of the 41st International Conference on Machine
  Learning (ICML)}, 2024.

\bibitem[Gu et~al.(2022)Gu, Goel, and R{\'e}]{gu2022s4}
Albert Gu, Karan Goel, and Christopher R{\'e}.
\newblock Efficiently modeling long sequences with structured state spaces.
\newblock In \emph{International Conference on Learning Representations
  (ICLR)}, 2022.

\bibitem[Smith et~al.(2023)Smith, Warrington, and Linderman]{smith2023s5}
Jimmy T.~H. Smith, Andrew Warrington, and Scott~W. Linderman.
\newblock Simplified state space layers for sequence modeling.
\newblock In \emph{International Conference on Learning Representations
  (ICLR)}, 2023.

\bibitem[Gu and Dao(2023)]{gu2023mamba}
Albert Gu and Tri Dao.
\newblock Mamba: Linear-time sequence modeling with selective state spaces.
\newblock \emph{arXiv preprint arXiv:2312.00752}, 2023.

\bibitem[Muca~Cirone et~al.(2024)Muca~Cirone, Orvieto, Walker, Salvi, and
  Lyons]{cirone2024selective}
Nicola Muca~Cirone, Antonio Orvieto, Benjamin Walker, Cristopher Salvi, and
  Terry Lyons.
\newblock Theoretical foundations of deep selective state-space models.
\newblock In \emph{Advances in Neural Information Processing Systems
  (NeurIPS)}, 2024.

\bibitem[Walker et~al.(2025)Walker, Yang, Muca~Cirone, Salvi, and
  Lyons]{walker2025slice}
Benjamin Walker, Lingyi Yang, Nicola Muca~Cirone, Cristopher Salvi, and Terry
  Lyons.
\newblock Structured linear {CDE}s: Maximally expressive and parallel-in-time
  sequence models.
\newblock In \emph{Advances in Neural Information Processing Systems
  (NeurIPS)}, 2025.

\bibitem[Tallec and Ollivier(2018)]{tallec2018warp}
Corentin Tallec and Yann Ollivier.
\newblock Can recurrent neural networks warp time?
\newblock In \emph{International Conference on Learning Representations
  (ICLR)}, 2018.

\bibitem[Marcus(1981)]{marcus1981}
Steven~I. Marcus.
\newblock Modeling and approximation of stochastic differential equations
  driven by semimartingales.
\newblock \emph{Stochastics}, 4\penalty0 (3):\penalty0 223--245, 1981.

\bibitem[Chevyrev and Friz(2019)]{chevyrev2019canonical}
Ilya Chevyrev and Peter~K. Friz.
\newblock Canonical {RDE}s and general semimartingales as rough paths.
\newblock \emph{The Annals of Probability}, 47\penalty0 (1):\penalty0 420--463,
  2019.

\bibitem[Cuchiero et~al.(2025)Cuchiero, Primavera, and
  Svaluto-Ferro]{cuchiero2025cadlag}
Christa Cuchiero, Francesca Primavera, and Sara Svaluto-Ferro.
\newblock Universal approximation theorems for continuous functions of
  c{\`a}dl{\`a}g paths and {L}{\'e}vy-type signature models.
\newblock \emph{Finance and Stochastics}, 29\penalty0 (2), 2025.

\bibitem[Chen(1957)]{chen1957}
Kuo-Tsai Chen.
\newblock Integration of paths, geometric invariants and a generalized
  {B}aker--{H}ausdorff formula.
\newblock \emph{Annals of Mathematics}, 65\penalty0 (1):\penalty0 163--178,
  1957.

\bibitem[Lyons(1998)]{lyons1998}
Terry~J. Lyons.
\newblock Differential equations driven by rough signals.
\newblock \emph{Revista Matem{\'a}tica Iberoamericana}, 14\penalty0
  (2):\penalty0 215--310, 1998.

\bibitem[Cartier and Foata(1969)]{cartier1969}
Pierre Cartier and Dominique Foata.
\newblock \emph{Probl{\`e}mes Combinatoires de Commutation et
  R{\'e}arrangements}, volume~85 of \emph{Lecture Notes in Mathematics}.
\newblock Springer, 1969.

\bibitem[Saito et~al.(2021)Saito, Nakamura, Yatabe, Koizumi, and
  Saruwatari]{saito2021sfi}
Koichi Saito, Tomohiko Nakamura, Kohei Yatabe, Yuma Koizumi, and Hiroshi
  Saruwatari.
\newblock Sampling-frequency-independent audio source separation using
  convolution layer based on impulse invariant method.
\newblock \emph{arXiv preprint arXiv:2105.04079}, 2021.

\bibitem[Zhang et~al.(2017)Zhang, Suda, Lai, and Chandra]{zhang2017helloedge}
Yundong Zhang, Naveen Suda, Liangzhen Lai, and Vikas Chandra.
\newblock Hello edge: Keyword spotting on microcontrollers.
\newblock \emph{arXiv preprint arXiv:1711.07128}, 2017.

\bibitem[Pazy(1983)]{pazy1983}
Amnon Pazy.
\newblock \emph{Semigroups of Linear Operators and Applications to Partial
  Differential Equations}.
\newblock Springer, 1983.

\end{thebibliography}
}

\clearpage
\appendix
\section{Proofs}
\label{app:proofs}

Throughout, $\|\cdot\|$ denotes the $\ell_\infty$ norm unless stated otherwise, and all states are assumed reachable.

\subsection{Proof of \cref{prop:homomorphism}}

(a) If the processor is partition invariant, comparing the partitions $\{a,c\}$ and $\{a,b,c\}$ gives $P(a,c]=P(b,c]\circ P(a,b]$. Conversely, if the composition law holds, induction on the number of cut points shows that every partition of $(a,c]$ yields $P(a,c]$.

(b) Let the events in $(a,b]$ be $e_\ell,\dots,e_m$ and those in $(b,c]$ be $e_{m+1},\dots,e_p$. Then $P(b,c]\circ P(a,b]$ contains the factor $\Phi_{t_{m+1}-b}\circ\Phi_{b-t_m}=\Phi_{t_{m+1}-t_m}$ by the semigroup property, which is the corresponding factor of $P(a,c]$; empty packets are handled in the same way. Because packets are half-open on the left and closed on the right, tie groups are never split.

(c) Let $E(\delta)$ be the map of an empty packet of length $\delta$; by time homogeneity it does not depend on the packet's position. By (a), $E(\delta+\delta')=E(\delta')E(\delta)$, with $E(0)=I$, and $E$ is continuous. A continuous one-parameter semigroup of $d\times d$ matrices is of the form $e^{\delta\Lambda}$~\cite{pazy1983}.\qed

\subsection{Proof of \cref{thm:commute} and its corollaries}

\emph{(i)$\Leftrightarrow$(iii)} is the definition. \emph{(ii)$\Leftrightarrow$(iii)}: since $(\delta,x)\mapsto\Phi_\delta(x)$ is continuous and $\Phi_0=\id$, and the jumps are continuous, $\lim_{\delta\downarrow0}J_b\Phi_\delta J_a(x)=J_bJ_a(x)$ and $\lim_{\delta\downarrow0}J_a\Phi_\delta J_b(x)=J_aJ_b(x)$. The two limits agree for every $x$ if and only if the jumps commute.

\emph{Global continuity.} Assume all pairs commute. For every ordering $\sigma$ of the event indices let $\mathcal{C}_\sigma=\{t:\ t_0\le t_{\sigma(1)}\le\dots\le t_{\sigma(m)}\le T\}$, a closed set. On $\mathcal{C}_\sigma$, \eqref{eq:state} with order $\sigma$ is a composition of continuous maps and hence continuous in $t$. On $\mathcal{C}_\sigma\cap\mathcal{C}_{\sigma'}$, any two events ordered differently by $\sigma$ and $\sigma'$ have equal time stamps. The two formulas therefore differ only by permutations inside tie groups, between whose members the drift is $\Phi_0=\id$, and a product of pairwise commuting maps does not depend on the order of its factors. So the formulas agree on every overlap. The sets $\mathcal{C}_\sigma$ form a finite closed cover, and the pasting lemma gives continuity of $x(T)$ on their union. Independence from the processing order of tie groups is the same observation.\qed

\emph{\Cref{cor:norescue}.} If $J_bJ_a(x)\ne J_aJ_b(x)$, the state immediately after the two events tends to different limits from the two sides of the crossing, and no single value at the tie equals both. When the subsequent dynamics is injective (as for flows of ODEs and for invertible linear drifts), the discontinuity persists at every later query. Group, Jacobi-type or Marcus-type rules change only the value at the tie, not the one-sided limits.

\emph{\Cref{cor:affine}.} $J_bJ_a(x)=G_bG_ax+G_bb_a+b_b$ and $J_aJ_b(x)=G_aG_bx+G_ab_b+b_a$. These agree for all $x$ if and only if $G_aG_b=G_bG_a$ and $G_bb_a+b_b=G_ab_b+b_a$, that is, $N_bb_a=N_ab_b$. If every $N_c$ reads only coordinates in a set $\mathcal{R}$ and writes only coordinates in a set $\mathcal{W}$ with $\mathcal{R}\cap\mathcal{W}=\emptyset$, then $N_aN_b=0=N_bN_a$, so $G_aG_b=I+N_a+N_b=G_bG_a$. If moreover every $b_c$ vanishes on $\mathcal{R}$, then $N_ab_b=0=N_bb_a$.\qed

\subsection{Proof of \cref{thm:timemediated}}

\emph{(a)} If every jump commutes with every $\Phi_\delta$, all drift factors in \eqref{eq:state} can be moved to the left. They combine to $\Phi_{T-t_0}$ by the semigroup property, and the remaining product of pairwise commuting jumps does not depend on their order. For the equivalence with $C_J\equiv0$: differentiating $J(\Phi_\delta(y))=\Phi_\delta(J(y))$ at $\delta=0$ gives $DJ(y)F(y)=F(J(y))$. Conversely, if $C_J\equiv0$, then $g(\delta)=J(\Phi_\delta(y))$ satisfies $g'(\delta)=DJ(\Phi_\delta y)F(\Phi_\delta y)=F(g(\delta))$, as does $h(\delta)=\Phi_\delta(J(y))$, and $g(0)=h(0)$; uniqueness of solutions for the locally Lipschitz field $F$ gives $g=h$.

\emph{(b)} Write $w=J_bJ_aY=J_aJ_bY$. Since the flow is $C^1$, $\Phi_\delta(z)=z+\delta F(z)+o(\delta)$ locally uniformly. Hence $J_b(\Phi_\delta(J_aY))=w+\delta\,p+o(\delta)$ with $p=DJ_b(J_aY)F(J_aY)$, and $J_a(\Phi_\delta(J_bY))=w+\delta\,p'+o(\delta)$ with $p'=DJ_a(J_bY)F(J_bY)$. Applying $\Phi_{T-t-\delta}$ and using differentiability of the flow in the state and continuity of $D\Phi$ in time, the difference is $\delta\,D\Phi_{T-t}(w)(p-p')+o(\delta)$. Finally $p=C_{J_b}(J_aY)+F(J_bJ_aY)$ and $p'=C_{J_a}(J_bY)+F(J_aJ_bY)$, and the $F$ terms cancel because $J_bJ_aY=J_aJ_bY$.

\emph{Affine formula~\eqref{eq:affineexact}.} Insert $\Phih_\delta=I+\int_0^\delta\Lh\Phih_\sigma d\sigma$ into $\Jh_b\Phih_\delta\Jh_a-\Jh_a\Phih_\delta\Jh_b$ and use $\Jh_b\Jh_a=\Jh_a\Jh_b$. The leading term follows from $\Jh_b\Lh\Jh_a-\Jh_a\Lh\Jh_b=[\Jh_b,\Lh]\Jh_a-[\Jh_a,\Lh]\Jh_b$, which uses the same commutation. For $J(y)=Gy+b$ and $F(y)=\Lambda y$, $C_J(y)=(G\Lambda-\Lambda G)y-\Lambda b$, which is the first block of $[\Jh,\Lh]\hat y$.\qed

\emph{\Cref{cor:tmclass}} combines \cref{thm:commute} with parts (a) and (b). \emph{\Cref{cor:resolution}}: swapping the types of the events at $t$ and $t+\delta$ is the same as moving each of the two events by $\delta$, an $\ell_1$ displacement of $2\delta$. \emph{\Cref{prop:binned}}: an event within $\varepsilon$ of an edge crosses it under a shift of at most $\varepsilon$, which moves one unit of count between two bins. For a time stamp uniform within its bin, the probability of lying within $\varepsilon\le\Delta/2$ of either edge is $2\varepsilon/\Delta$. The TM statement follows from \cref{thm:lipschitz}.

\subsection{Proof of \cref{prop:closedform} and \cref{lem:kernel}}

Between events each coordinate pair $(u_k,s_k)$ obeys a $2\times2$ lower-triangular linear system whose response in $s_k$ to a unit initial value of $u_k$ is $K_k(\tau)$. Jumps add $\alpha_c$ to $u$ and never modify $s$, so by superposition $u(T)$ and $s(T)$ are as stated; an event at time $T$ contributes $K(0)=0$ to $s(T)$. At event $j$, the jump reads $s(t_j)$, which is the sum of $K(t_j-t_i)\alpha_{c_i}$ over the events processed before $j$. Events in $j$'s tie group contribute $K(0)=0$, and later events contribute nothing, so the sum can be taken over all $i$ with the convention $K(\tau)=0$ for $\tau\le0$. Superposing the decayed jumps of $v$ gives the formula.

For the lemma, $K_k(\tau)$ is the value at time $\tau$ of $s$ solving $\dot s=\lambda_s(\rho u-s)$, $s(0)=0$, with $u(r)=e^{-\lambda_ur}\in(0,1]$. Hence $0\le K_k(\tau)=\int_0^\tau\lambda_se^{-\lambda_s(\tau-r)}\rho u(r)dr\le\rho$. Moreover $K_k'(\tau)=\lambda_s\rho\,[f(\lambda_u)-f(\lambda_s)]/(\lambda_u-\lambda_s)$ with $f(\lambda)=\lambda e^{-\lambda\tau}$, and $|f'(\lambda)|=|e^{-\lambda\tau}(1-\lambda\tau)|\le1$, so $|K_k'|\le\lambda_s\rho$ by the mean value theorem; the case $\lambda_u=\lambda_s$ is the limit. At $\tau=0$, $K$ is continuous with one-sided derivatives $0$ and $\lambda_s\rho$, so $K$ is Lipschitz on $\R$.\qed

\subsection{Proof of \cref{thm:lipschitz}}

The closed form of \cref{prop:closedform} holds for every configuration, whatever the order of events, so it suffices to bound the Lipschitz constant of each term in each time stamp. For $u$, $|e^{-\lambda_u(T-t)}-e^{-\lambda_u(T-t')}|\le\lambda_u|t-t'|$ for $t,t'\le T$, which gives $\lambda_uA$. For $s$, \cref{lem:kernel} gives $\lambda_s\rho A$. For $v$, the term indexed by $(j,i)$ is $e^{-\lambda_v(T-t_j)}g_{c_j}R_{c_j}K(t_j-t_i)\alpha_{c_i}$, whose norm is at most $g_{\max}\|R\|_\infty\rho A$. Its Lipschitz constant in $t_j$ is at most $g_{\max}\|R\|_\infty A\rho(\lambda_v+\lambda_s)$, and in $t_i$ at most $g_{\max}\|R\|_\infty A\rho\lambda_s$. A given time stamp appears as $j$ in at most $n$ terms and as $i$ in at most $n$ terms. Summing gives the third entry of $C$. Gates depend only on intrinsic marks, not on time.\qed

\subsection{Proof of \cref{thm:separation}}

\begin{lemma}[Collocation]
\label{lem:collocation}
If functions $\phi_1,\dots,\phi_m$ on a set $E$ are linearly independent, there are points $e_1,\dots,e_m\in E$ with $\det[\phi_i(e_j)]\ne0$.
\end{lemma}
\begin{proof}
By induction on $m$. Given points for $\phi_1,\dots,\phi_{m-1}$ with nonzero determinant, expand $e\mapsto\det[\phi_i(e_j)]_{i,j\le m}$, with $e_m=e$, along its last column. The result is a linear combination $\sum_i\gamma_i\phi_i(e)$ whose coefficient $\gamma_m$ is $\pm$ the nonzero minor, so the combination is not identically zero, and some $e_m$ makes it nonzero.
\end{proof}

\emph{Additivity.} For translation jumps and linear drift, $x(T)=e^{(T-t_0)\Lambda}x_0+\sum_{t_i\le T}e^{(T-t_i)\Lambda}b_{c_i}$. Stacking the $q$ queries, the collected state is $X=X_0+\sum_ih_{c_i}(t_i)$ with $h_c(t)=\bigl(\mathbf{1}[t\le T_\ell]\,e^{(T_\ell-t)\Lambda}b_c\bigr)_{\ell=1}^q\in\R^{qd}$.

\emph{Choice of points.} Fix an open interval $I$ that contains no query time, and $t^\ast\in I$. On $I$ each component of $h_a$ is real analytic (or identically zero). Let $f_1,\dots,f_r$ ($r\le qd$) be a basis of the span $V$ of the components of $h_a'$ on $I$, and let $g(t)=\frac{d}{dt}\kappa(t^\ast-t)$, which equals $-\kappa'(t^\ast-t)$ for $t<t^\ast$ and $0$ for $t>t^\ast$. Since $\kappa'(0^+)=1/20$, $g$ has a jump at $t^\ast$, whereas every element of $V$ is continuous on $I$; hence $f_1,\dots,f_r,g$ are linearly independent on $I\setminus\{t^\ast\}$. By \cref{lem:collocation} there are distinct points $s_1,\dots,s_{r+1}\in I\setminus\{t^\ast\}$ whose $(r+1)\times(r+1)$ collocation matrix is invertible.

\emph{A fiber of constant state.} Write $f_m=\beta_m^\top h_a'$ and $F_m=\beta_m^\top h_a$, and let $\psi_m(\tau)=\sum_{k=1}^{r+1}F_m(\tau_k)$ for $\tau\in I^{r+1}$. Every component of $h_a'$ is a combination $\sum_m\gamma_{pm}f_m$, so $(h_a)_p-\sum_m\gamma_{pm}F_m$ is constant on $I$. Hence the $a$-part of the state, $\sum_kh_a(\tau_k)$, is an affine function of $\psi(\tau)$, and level sets of $\psi$ are level sets of the state. The Jacobian $D\psi(s)=[f_m(s_k)]$ has rank $r$, being $r$ rows of an invertible matrix. By the implicit function theorem, $\psi^{-1}(\psi(s))$ is near $s$ a $C^1$ curve $\gamma(\theta)$ with $\gamma(0)=s$ and nonzero tangent $w$ spanning $\ker D\psi(s)$. Place the $a$-events at $\gamma(\theta)$ and $N$ $b$-events at $t^\ast$. Along $\gamma$ the collected state, and therefore the output of $F$, is constant.

\emph{The target changes.} On these streams $\yord=N\,G(\gamma(\theta))$ with $G(\tau)=\sum_k\kappa(t^\ast-\tau_k)$, whose gradient at $s$ is $(g(s_k))_k$. If $\nabla G(s)\cdot w=0$, then $\nabla G(s)$ lies in $(\ker D\psi(s))^\perp$, the row space of $D\psi(s)$, and the collocation matrix of $(f_1,\dots,f_r,g)$ would be singular. Hence $G$ is strictly monotone along $\gamma$ near $\theta=0$. Choose $\theta_1\ne\theta_2$ and set $c=|G(\gamma(\theta_1))-G(\gamma(\theta_2))|>0$, which does not depend on $N$. The two streams give the same output, and their targets differ by $Nc$.

\emph{Recency features.} Choose the points in a subinterval $I'\ni t^\ast$ with $\overline{I'}\subset I$, and add one ``sentinel'' $a$-event at a fixed time in $I$ to the right of $I'$. Every query after $I$ then sees the sentinel as the most recent $a$-event and $t^\ast$ as the most recent $b$-event in both streams, and queries before $I$ see neither. The sentinel adds the same amount to the state in both streams and nothing to $\yord$, since it follows all $b$-events. The streams have at most $qd+2$ $a$-events.\qed

\subsection{Proof of \cref{prop:tmyord}}

With $\lambda_s=\lambda_u=\lambda=1/20$ and $\rho=1$, $K(\tau)=\lambda\tau e^{-\lambda\tau}=\kappa(\tau)$. With the stated parameters, $u$ and $s$ count only $a$-events and only $b$-events write $v$, so \cref{prop:closedform} gives the formula. Then $\yord-v(T)=\sum_j\bigl(1-e^{-\lambda_v(T-t_j)}\bigr)\sum_i\kappa(t_j-t_i)$, which lies between $0$ and $(1-e^{-\lambda_vW})\yord$ because $\kappa\ge0$ and $T-t_j\le W$.\qed

\subsection{Proof of \cref{prop:height}}

\emph{Upper bounds.} Marks of the first layer are intrinsic, so they have height $0$. If the marks of a layer have height at most $h$, then every additive coordinate $\sum_{t_i\le t}w(t-t_i)(b_{c_i}+Bz_i)$ has height at most $h+1$. For a TM layer, $u$ and $s$ have height at most $h+1$. The $v$-summand $g(z_j)R\,s(t_j)$ is a product of features of heights at most $h$ and $h+1$, so it has height at most $h+1$, and the outer sum gives $h+2$. Induction over layers gives $k$ and $2k$.

\emph{Attainability.} Let $C_{2\ell}(t)$ denote the chain functional over the types $a_1,\dots,a_{2\ell}$ restricted to $t_{i_{2\ell}}\le t$. The first layer produces $C_2$ in the limit $\lambda_v\to0$ (\cref{prop:tmyord}). Suppose layer $\ell$ produces $C_{2\ell}$ in its $v$. Layer $\ell+1$ receives an event at each input time, with a type that identifies the input type and mark $z=v^{(\ell)}$, evaluated after the tie group. Let events of type $a_{2\ell+1}$ write $u\leftarrow u+z$, set $\lambda_u=\lambda_s=1/20$ and $\rho=1$, and let events of type $a_{2\ell+2}$ read $s$ into $v$. By \cref{prop:closedform},
$v^{(\ell+1)}(T)=\sum_{j\in a_{2\ell+2}}e^{-\lambda_v(T-t_j)}\sum_{i\in a_{2\ell+1}}\kappa(t_j-t_i)\,C_{2\ell}(t_i)$,
which tends to $C_{2\ell+2}(T)$ as $\lambda_v\to0$. The link $i_{2\ell}\le i_{2\ell+1}$ is non-strict because the mark is evaluated after the tie group.\qed

\subsection{Proof of \cref{prop:tmk}}

Each jump's linear part maps some $s_j$ into $v_j$ only, and no jump reads $u$ or any $v_j$; the constant parts touch only $u$. By \cref{cor:affine} all jumps commute, and \cref{thm:commute,prop:homomorphism} give the three properties. For the closed form, superposition applies as in \cref{prop:closedform}. A jump $\Delta$ in $v_j$ at time $t'$ contributes $K_{j+1}(t-t')\Delta$ to $s_{j+1}$, where $K_{j+1}(\tau)=\Lambda_{s_{j+1}}P_{j+1}\varphi(\Lambda_{s_{j+1}},\Lambda_{v_j},\tau)$ for $\tau>0$ and $0$ otherwise; $K_1$ is the kernel of \cref{prop:closedform}. Iterating, $s_j(t)=\sum_{i}K_j(t-t_i)\,g_{c_i,j-1}R_{c_i,j-1}s_{j-1}(t_i)$, and the formula for $v_{k-1}(T)$ follows. All inequalities are strict because $K_j(\tau)=0$ for $\tau\le0$.\qed

\subsection{Proof of \cref{prop:stacking}}

By the iteration in the proof of \cref{prop:tmk}, every drift-mediated coordinate at time $t$ is a finite sum of terms $K(t-t_i)\,\xi_i$. Each kernel $K$ is continuous on $\R$ and vanishes for $\tau\le0$, and each $\xi_i$ is a continuous function of earlier drift-mediated coordinates. By induction these coordinates are continuous functions of $(t,t_1,\dots,t_n)$ and have no jumps at event times. Hence the marks passed to layer $\ell+1$ are continuous in the time stamps, and at a tie group they take the same value whichever tied events have already been processed. The state of layer $\ell+1$ is given by \eqref{eq:state} with jump maps $J_{c,z}$. On each closed set of fixed order it is continuous in the time stamps and in the marks. On overlaps the formulas agree, because jumps commute for all mark values and marks do not depend on the order within ties. The pasting lemma completes the proof, and partition invariance follows from exact processing. If instead a mark reads $u$ or $v$, the mark of an $a$-event computed at a tie with a $b$-event includes $b$'s jump, whereas just before the crossing it does not. The mark then changes by a full jump, and the next layer's input is discontinuous for generic parameters.\qed

\subsection{Proof of \cref{prop:sampling}}

Only finitely many levels lie in the range of $x$, and each has finitely many transversal crossings. Hence there are $\epsilon_0,\eta>0$ such that $|x-L|\ge\eta$ outside the $\epsilon_0$-neighbourhoods of the crossings of each level $L$, and $|x'|\ge\mu/2$ inside them. The linear interpolant satisfies $|\hat x-x|\le Mh^2/8$. For $Mh^2/8<\eta$, $\hat x$ cannot cross a level outside the neighbourhoods. Inside, the slope of $\hat x$ on any sampling interval equals $x'(\xi)$ for some $\xi$ in that interval, which has the sign of $x'$ at the crossing and magnitude at least $\mu-Mh$. So for $h$ small, $\hat x$ crosses each level exactly once near each true crossing, with the same sign. At a true crossing $t^\ast$ of level $L$, $|\hat x(t^\ast)-L|\le Mh^2/8$, and the interpolant's crossing $\hat t$ satisfies $|\hat t-t^\ast|\le (Mh^2/8)/(\mu-Mh)$. The state bound follows from \cref{thm:lipschitz} applied to the $\ell_1$ displacement of the $n$ time stamps.\qed

\subsection{Proof of \cref{prop:compensator}}

Between events the state evolves by the linear drift~\eqref{eq:closeddrift}. Each coordinate is a combination of $e^{-\lambda\tau}$ terms and, through $\varphi$, either differences of exponentials or, for equal rates, $\tau e^{-\lambda\tau}$. Therefore the entries of $h$ lie in the span of $\tau^pe^{-\gamma\tau}$ with $p\le1$, and those of $hh^\top$ in the span with $p\le2$; the integrals $\int_0^\Delta\tau^pe^{-\gamma\tau}d\tau$ are elementary. Positivity follows from $Q_k\succeq0$ and $\mu_k\ge0$. The log-likelihood $\sum_i\log\lambda_{k_i}(t_i)-\sum_k\int\lambda_k$ is then exact. The same holds for TM-$k$ states, whose coordinates are exponential polynomials of higher degree.\qed

\subsection{Justification of \cref{tab:taxonomy}}
\label{app:taxonomy}

\emph{Binned networks.} The binned input changes when an event crosses a bin edge, under arbitrarily small retimings (\cref{prop:binned}) and under a change of bin phase, so (P) and (C) fail for any non-constant network. Within a bin, counts are sums, so (T) holds. Recurrence across bins forms interactions between events in different bins, at resolution $\Delta$.

\emph{Additive event SSMs.} Translations commute, so (T) and (C) hold by \cref{thm:commute}; exact processing gives (P); and a single layer cannot form ordered pairs (\cref{thm:separation}).

\emph{Selective affine updates.} For $v\leftarrow v+g_cR_cu$ with $u\leftarrow u+\alpha_c$, \cref{cor:affine} gives non-commutation for generic parameters. The measured tie and swap differences in \cref{tab:checks} confirm it. The update reads $u$, which contains earlier events, so ordered pairs are formed.

\emph{Gated nonlinear updates.} A GRU whose update gate is saturated reduces to $h\leftarrow\tanh(Wh+Ux_c)$. With scalar $W=1$, $Ux_a=1$ and $Ux_b=-1$, we get $J_aJ_b(0)=\tanh(1-\tanh1)\approx0.234$ but $J_bJ_a(0)=\tanh(\tanh1-1)\approx-0.234$. The updates do not commute, so (T) and (C) fail by \cref{thm:commute}. Exact integration of the drift gives (P).

\emph{Online log-signature models.} Interval composition gives (P)~\cite{walker2026faithful}. That formalism assumes strictly increasing observation times, and simultaneous observations in several channels are encoded as a single event, so (T) is excluded by assumption. Continuity is established within a fixed pattern of events. When the generators of two channels do not commute, as in the experiments of~\cite{walker2026faithful}, \cref{cor:norescue} implies a discontinuity when observations of the two channels cross in time. Level-two terms form ordered pairs.

\emph{TM.} \Cref{cor:affine,thm:commute,prop:closedform,thm:lipschitz}.

\section{Implementation and Training Details}
\label{app:details}

\paragraph{Stable evaluation of $\varphi$.} With $z=(b-a)\Delta$: for $z\ge0$, $\varphi(a,b,\Delta)=\Delta e^{-a\Delta}E(z)$; for $z<0$, $\varphi(a,b,\Delta)=\Delta e^{-b\Delta}E(-z)$; here $E(z)=-\mathrm{expm1}(-z)/z$, evaluated by its Taylor series $1-z/2+z^2/6-z^3/24$ for $|z|<10^{-3}$. This expression is accurate when the two rates nearly coincide and for long gaps, and it gives $\varphi(\cdot,\cdot,0)=0$.

\paragraph{Time and windows.} Time stamps are kept as 64-bit integers in microseconds; millisecond-floored stamps are never used as input. Events with equal stamps form a tie group. On DVS128 Gesture, training uses five cold-start crops of 800\,ms per clip, starting at 0, 50, 100, 150 and 200\,ms. Queries are placed every 50\,ms, and the loss weights the 13 prefixes from 200 to 800\,ms equally. Each window starts from a zero state with empty pixel tables.

\paragraph{Raw query features.} For each site and polarity: the mean over the site's pixels of $\exp(-(T-t^{\mathrm{last}}_{\mathrm{pix}})/\tau)$ for $\tau\in\{5,20,100\}$\,ms (zero for pixels without events), and $\log(1+c)$ of a site-level count that decays with $\tau=20$\,ms. These eight features are shared by all arms; the \zero arm sees only these features, with its site state set to zero.

\paragraph{Parameters.} Time constants are parameterized as $\tau=\exp(\theta)$, initialized log-uniformly and clamped to $[2,2000]$\,ms; $\rho$ is initialized to 1. Marks are the two coordinates of the pixel (or channel) within its site.

\paragraph{Optimization and selection (DVS128 Gesture).} AdamW with learning rate $10^{-3}$, weight decay $10^{-4}$, batch size 16, gradient clipping at 1, label smoothing 0.05, and an exponential moving average of the weights. The selected checkpoint has the lowest calibration cross-entropy at 800\,ms. Training runs for at least 120 epochs and stops after 50 epochs without improvement. Augmentation (V2), applied to raw events before grouping and in this order: spatial scaling in $[0.85,1.15]$; translation by up to $\pm12$ pixels, discarding events that leave the sensor; time stretching by a factor in $[0.8,1.2]$; random deletion with probability drawn from $[0,0.2]$; then rounding to microseconds, sorting and regrouping.

\paragraph{SHD.} The 700 channels are grouped into sites of 8 adjacent channels, with a one-dimensional 3-site neighbourhood and 6 letters; the mark is the channel's position within its site. The readout takes the site states at 10 equally spaced query times and applies a small MLP and a time-averaged classifier. Configuration A adds time stretching in $[0.8,1.2]$, a random whole-sample channel shift of up to $\pm10$ channels and random event deletion with probability in $[0,0.2]$. Within each run, the checkpoint, learning-rate schedule and early stopping are driven by the validation cross-entropy at 1000\,ms; configurations are compared by the mean accuracy over the 10 query times.

\paragraph{Speech Commands.} We use the 35-class task with the official validation list (9,981 clips). Waveforms are converted to events by a fixed encoder shared by all arms. The DS-CNN baseline~\cite{zhang2017helloedge} is trained on the same split with its frame-based front end. The \zero arm on this dataset was stopped manually after 16 epochs; its best validation accuracy was 6.87\% (35 classes).

\paragraph{PLIF baselines.} PLIF networks are implemented in SpikingJelly~\cite{fang2021plif,fang2023spikingjelly}. Recipe (i) trains on complete clips converted to frames. Recipe (ii) first truncates each clip to a prefix and then applies the baseline's own frame conversion, with our multi-prefix loss. The augmented variant applies the same event augmentation and a frame-level count gain.

\paragraph{Numerical checks.} The verification code implements a single site independently of the training code, in FP64, with three $u/s$ channels, two $v$ channels and two event types. Time constants are drawn in $[2,200]$\,ms for $u$ and $s$ and in $[20,2000]$\,ms for $v$; $\rho\in[0.5,1.5]$; $\alpha,\beta$ and $R$ (scaled by $1/\sqrt3$) are Gaussian, with gates $1+\tanh(\cdot)$ of a Gaussian. For the fiber check: $\lambda_u=(1/5,1/60)$, $\lambda_s=(1/15,1/200)$, $\lambda_v=1/35$, $\alpha_a=(1,0.7)$ and $\beta_a=0.9$ (all per ms), a single query at 200\,ms and 50 $b$-events at 120\,ms. One $a$-event moves from 114 to 117\,ms while five others, initially at 20, 55, 95, 160 and 190\,ms, follow the constant-state curve computed by damped Newton continuation.

\section{SHD Site-Size and Capacity Study}
\label{app:shd}

\Cref{tab:shd} lists TM variants on the SHD validation split. Each variant changes one factor relative to configuration A. Accuracy peaks at 8 channels per site. Fourfold wider states, more aggressive channel-shift augmentation, and channel-position letters did not improve on configuration A.

\begin{table}[h]
\centering
\caption{SHD validation accuracy (\%) of TM variants.}
\label{tab:shd}
\small
\setlength{\tabcolsep}{3pt}
\begin{tabular}{@{}>{\raggedright\arraybackslash}p{0.56\columnwidth}cc@{}}
\toprule
Variant & 10-point mean & 1000\,ms \\
\midrule
Base (no augmentation, 8 channels per site) & 73.25 & 76.45 \\
A: augmented, 8 channels per site & \textbf{81.80} & \textbf{87.40} \\
A with 2 channels per site & 63.89 & 63.43 \\
A with 4 channels per site (B) & 75.83 & 82.27 \\
B with channel-position letters (12 letters) & 71.83 & 75.07 \\
A with 16 channels per site & 74.61 & 77.15 \\
A with $4\times$ state ($d_u=d_s=32$, $d_v=64$) & 80.42 & 85.18 \\
A with channel shift up to $\pm16$ & 76.83 & 81.16 \\
\bottomrule
\end{tabular}
\end{table}

\end{document}